\documentclass{article}

\usepackage[eandd, preprint]{neurips_2026}

\usepackage[utf8]{inputenc} 
\usepackage[T1]{fontenc}    
\usepackage{hyperref}       
\usepackage{url}            
\usepackage{booktabs}       
\usepackage{amsfonts}       
\usepackage{nicefrac}       
\usepackage{microtype}      
\usepackage{xcolor}         

\usepackage{booktabs}
\usepackage{makecell}
\usepackage{amssymb} 
\usepackage{graphicx}
\usepackage{wrapfig}
\usepackage{caption}
\usepackage{tabularx}
\usepackage{array}
\usepackage{amsmath}
\usepackage{tcolorbox}
\usepackage{enumitem}
\usepackage{subcaption}
\usepackage{listings}
\usepackage{multirow}
\usepackage{longtable}
\usepackage{fontawesome5}

\tcbset{
    colback=gray!5,
    colframe=gray!50,
    boxrule=0.5pt,
    arc=2pt,
    left=6pt,
    right=6pt,
    top=4pt,
    bottom=4pt
}

\setcitestyle{numbers,square}

\definecolor{takeawaysColor}{HTML}{233454} 

\definecolor{tred}{RGB}{227, 11, 92}

\newcommand{\myparagraph}[1]{\textbf{#1}}

\newcommand{\hficon}{%
  \raisebox{-0.2\height}{%
    \includegraphics[height=1.4em]{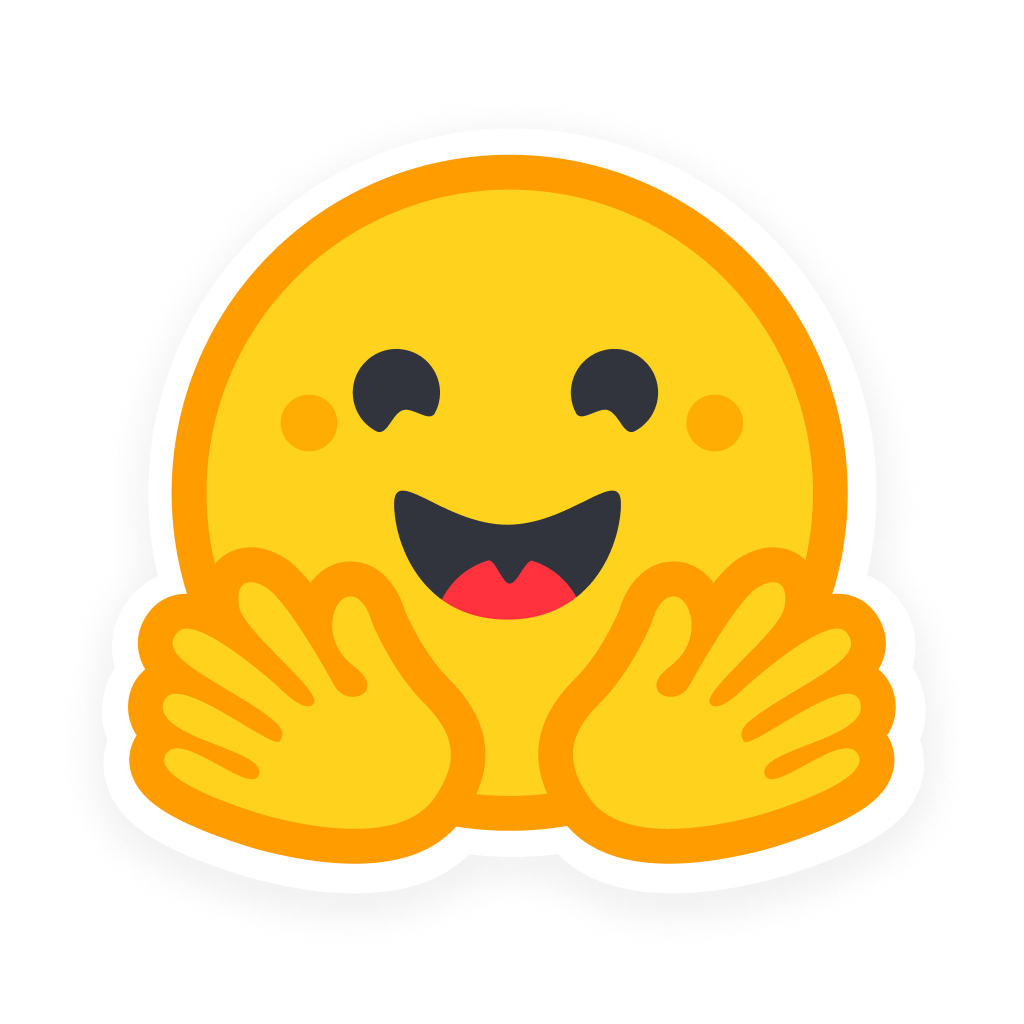}%
  }%
}

\title{MedKIT: Evaluating Knowledge Integration and Generalization in Large Language Models}

\author{%
  Lukas Thede$^{1,2,3}$ \And
  Yash Kumar Atri$^{4}$ \And
  David Chen$^{5}$ \AND
  Danielle Bitterman$^{6}$ \And
  Matthias Bethge$^{1}$ \And
  Tom Hartvigsen$^{4}$ \And
  Zeynep Akata$^{2,3,7}$ \\[1ex]
  $^{1}$University of T\"ubingen, T\"ubingen AI Center \quad
  $^{2}$Helmholtz Munich \\
  $^{3}$Munich Center for Machine Learning (MCML) \quad
  $^{4}$University of Virginia \\
  $^{5}$University of British Columbia \quad
  $^{6}$Harvard Medical School \quad
  $^{7}$Technical University of Munich \\
  \texttt{lukas.thede@uni-tuebingen.de}
}

\begin{document}

\maketitle

\begin{abstract}
Constantly evolving real-world knowledge necessitates models to be updated continuously. Especially in medicine, as clinical evidence changes over time, outdated knowledge can pose safety risks. Existing evaluations of knowledge integration focus on factual recall, offering limited insight into whether newly integrated knowledge is actually usable.
Our benchmark \textbf{MedKIT} (Medical Knowledge Integration and Transfer) provides a granular evaluation of how models integrate and apply knowledge under realistic sequences of clinical updates. Each instance corresponds to a factual update derived from clinical evidence, paired with targeted probes that assess transfer across lexical variation, relational transformations, compositional reasoning, and open-ended operationalization, as well as locality tests for knowledge preservation.
Using MedKIT, we conduct a large-scale empirical study of 12 knowledge integration strategies across 5 diverse models, including both general-purpose and medical LLMs. Our results reveal a consistent gap between recall and usable knowledge: while most methods achieve strong gains on the original update task and under lexical variation, relational generalization is limited, and no method yields meaningful improvements on compositional or operational tasks. 
These findings highlight a fundamental challenge in knowledge integration and position MedKIT as a testbed for developing methods that make newly integrated knowledge more consistently usable across tasks and contexts.
\end{abstract}

\begin{center}
\small
\href{https://huggingface.co/datasets/bethgelab/MedKIT}{%
    \hficon\ \textbf{MedKIT Dataset}%
}
\hspace{1.5em}
\href{https://github.com/bethgelab/MedKIT}{%
    \scalebox{1.3}{\faGithub}\ \textbf{Code}%
}
\end{center}

\section{Introduction}
While our world knowledge evolves continuously, static large language model (LLM) weights may generate outdated or incorrect answers. High-stakes domains like medicine provide a clear example of this challenge. Oncology provides a particularly relevant setting: cancer affects millions of patients worldwide~\citep{who2026cancer}, while new clinical evidence continuously updates our understanding of the relative efficacy of available treatments. Keeping LLMs aligned with such evolving evidence requires mechanisms to \emph{integrate new knowledge into already deployed models}.

A broad range of approaches has been proposed to update LLMs' knowledge, including knowledge editing \citep{meng2023rome, Meng2022memit}, continual post-training \citep{hu2021lora}, and retrieval-augmented generation (RAG) \citep{lewis2020rag1}. However, successfully integrating new knowledge is not merely a matter of storing it. In practice, models must be able to \emph{use newly acquired knowledge consistently across tasks and contexts}. This requires generalization beyond the original update formulation, including robustness to paraphrasing, relational transformations, and more complex reasoning settings. The gap between factual recall and usable knowledge is therefore central to evaluating knowledge integration.

\begin{figure*}[t]
    \centering
    \includegraphics[width=\textwidth]{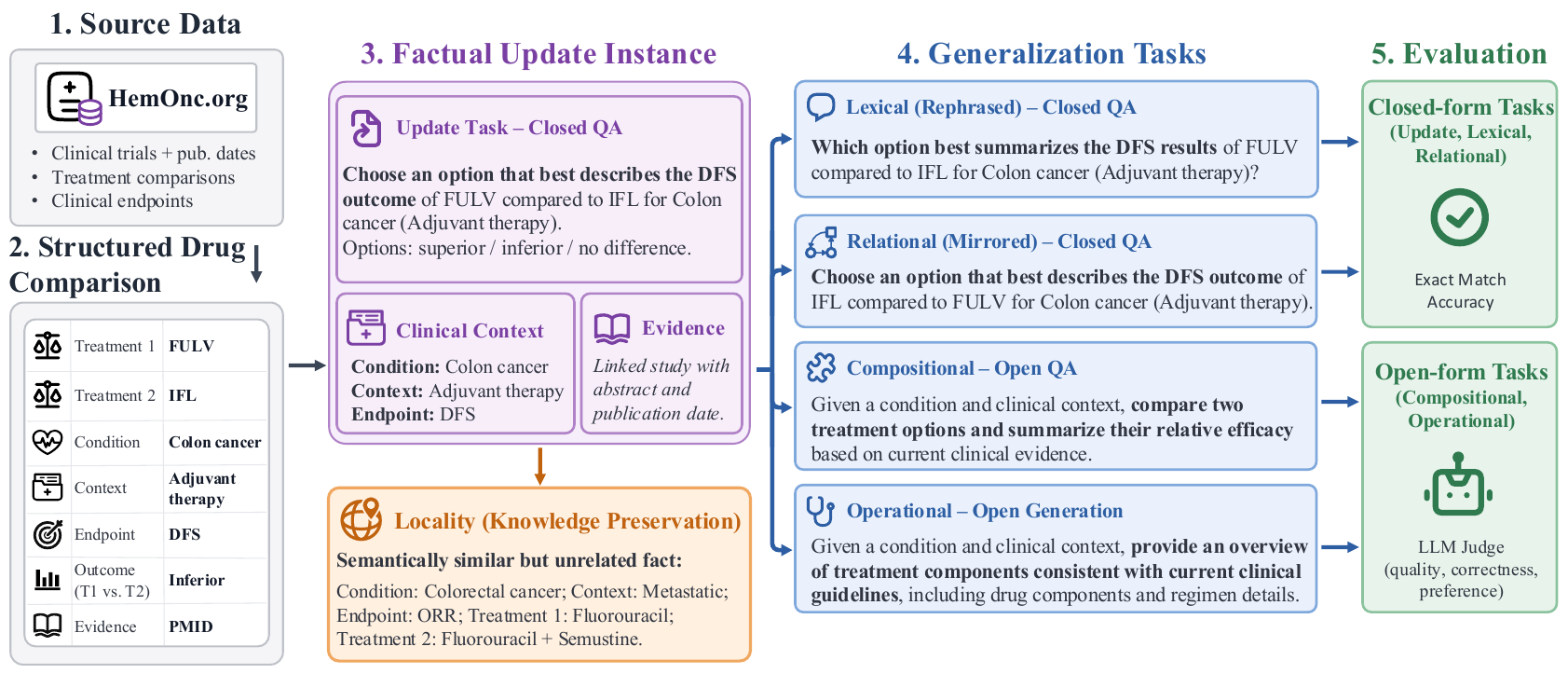}
    \caption{
    \textbf{MedKIT construction pipeline and examples.} Structured clinical comparisons are converted into factual updates and multiple task variants probing lexical, relational, compositional, and operational generalization, alongside locality tests for knowledge preservation.  
    }
    \label{fig:benchmark_construction}
    \vspace{-10pt}
\end{figure*}   

Current benchmarks for factual knowledge integration (e.g. \citep{levy2017zsre, meng2023rome}) provide limited insight into how well newly acquired knowledge generalizes (see Table~\ref{tab:editing_benchmarks}).
While some settings probe generalization through paraphrased or multi-hop questions, these evaluations test only isolated forms of transfer and do not capture how consistently updated knowledge is used across diverse task formulations. As a result, they provide little insight into how far factual updates extend beyond their original query. 
Moreover, many benchmarks rely on static or synthetic evaluation settings that lack a natural temporal structure, limiting their ability to reflect how knowledge evolves in real-world applications.

To address these limitations, we introduce \textbf{MedKIT}, a benchmark for Medical Knowledge Integration and Transfer that evaluates how models integrate and apply knowledge under realistic sequences of clinical updates. We focus on oncology, where clinical trials provide temporally grounded evidence on the comparative efficacy of treatment regimens that evolves as new study results become available. MedKIT uses these treatment comparisons as atomic knowledge updates and pairs each update with \textbf{7 targeted probes} that enable a \emph{granular evaluation of generalization} across lexical variation, relational transformations, compositional reasoning, and open-ended application, alongside locality probes for knowledge preservation.
For instance, given an update stating that FULV is inferior to IFL for colon cancer in the adjuvant setting, MedKIT probes whether models correctly handle inverted comparisons, open-ended reasoning tasks, and recommendations consistent with the underlying comparison (see Figure~\ref{fig:benchmark_construction}).
The benchmark comprises \textbf{6,196 canonical updates} extracted from clinical studies published between 1960 and 2026, enabling evaluation of whether newly integrated knowledge \emph{transfers across task formulations}.

Using MedKIT, we conduct a large-scale empirical study of knowledge integration strategies. We evaluate 12 approaches, including knowledge editing, continual post-training, and retrieval-based methods, across 5 models from both general and medical domains. This setup enables a systematic and fine-grained analysis of knowledge integration and generalization. 

Our results reveal that, while most methods achieve strong performance on the original update task and lexical variations, relational generalization remains a significant challenge. On compositional and operational generalization tasks, none of the methods achieved meaningful improvements over pre-update performance. Maintaining performance on past updates and avoiding interference with unrelated knowledge or general capabilities also emerge as additional difficulties.
The results highlight the current frontier in the generalization of integrated factual updates. 

Our work makes the following contributions: (1) We introduce \textbf{MedKIT}, a benchmark grounded in real-world clinical evidence to evaluate knowledge integration in realistic, evolving settings.
(2) We introduce a \textbf{multi-level generalization framework} and operationalize it through targeted evaluation probes, enabling a fine-grained analysis of how newly integrated knowledge transfers across tasks and contexts.
(3) We conduct a \textbf{large-scale empirical study} of knowledge integration strategies, evaluating knowledge editing, continual post-training, and retrieval-based methods across diverse model families, revealing substantial gaps between recall and consistent use across task formulations.

\section{Related Work}
\textbf{Knowledge Editing and Factual Knowledge Benchmarks.}
The integration of factual updates into large language models without retraining from scratch, while preserving unrelated knowledge, has been widely studied within the knowledge-editing paradigm. Early benchmarks such as CounterFact \citep{meng2023rome} and ZSRE \citep{levy2017zsre} focus on recalling edited facts and their paraphrases, emphasizing single-hop recall and locality. Subsequent work extends this setting to evaluate whether edits propagate to dependent facts (MQuAKE \citep{zhong2024mquake}), enable document-level reasoning (DocTER \citep{wu2025docter}), or capture dependencies between facts (UniEdit \citep{chen2025uniedit}; ThinkEval \citep{baser2026thinkeval}). Domain-specific datasets, such as MedEditBench \citep{chen2026mededitbench}, MedMKEB \citep{xu2025medmkeb}, and MedCF/MedFE \citep{xu2024medlasa}, increase realism by grounding evaluation in clinical settings and multimodal inputs. In parallel, lifelong editing benchmarks \citep{thede2025wikibigedit, cheng2025sLKE, cao2025mrlf} study sequential updates, with a primary focus on scalability and stability.

\begin{wraptable}{r}{0.62\textwidth}
    \centering
    \vspace{-10pt}
\centering
\footnotesize
\renewcommand{\arraystretch}{1.1}
\caption{\textbf{Comparing factual knowledge integration benchmarks.} Prior work emphasizes factual recall and isolated generalization settings, with limited coverage of multi-level transfer, open-ended tasks, and realistic update scenarios. (G: General, M: Medical, L: Lexical, R: Relational, C: Compositional, O: Operational)}
\resizebox{0.55\textwidth}{!}{
\begin{tabular}{lcccccc}
\toprule
\textbf{Benchmark}
& \textbf{Domain}
& \textbf{Input}
& \multicolumn{4}{c}{\textbf{Generalization}} \\
\cmidrule(lr){4-7}
&  & 
& \textbf{L}
& \textbf{R}
& \textbf{C}
& \textbf{O} \\
\midrule

CounterFact \citep{meng2023rome}
& G & Triples & \checkmark & - & - & - \\

ZSRE \citep{levy2017zsre}
& G & QA & \checkmark & - & - & - \\

WikiBigEdit \citep{thede2025wikibigedit}
& G & QA / Wiki facts & \checkmark & \checkmark & \checkmark & - \\

MQuAKE \citep{zhong2024mquake}
& G & QA (multi-hop) & - & \checkmark & - & - \\

DocTER \citep{wu2025docter}
& G & Documents & - & - & - & - \\

ThinkEval \citep{baser2026thinkeval}
& G & Graph & - & - & - & - \\

UniEdit \citep{chen2025uniedit}
& G & Graph & - & - & - & - \\

MedEditBench \citep{chen2026mededitbench}
& M & QA & - & - & \checkmark & - \\

MedMKEB \citep{xu2025medmkeb}
& M & Multi-modal & - & - & - & - \\

MedCF / MedFE \citep{xu2024medlasa}
& M & QA / Explanation 
& \checkmark & - & - & - \\

TiEBe \citep{almeida2025tiebe}
& G & QA (temporal events)
& \checkmark 
& - 
& - 
& - \\

RECALL \citep{liu2023recall}
& G & QA + external knowledge
& \checkmark 
& - 
& - 
& - \\

\midrule
\textbf{MedKIT (Ours)}
& \textbf{M} & \textbf{Clinical trials}
& \checkmark 
& \checkmark 
& \checkmark 
& \checkmark \\
\bottomrule
\end{tabular}
}
\label{tab:editing_benchmarks}
\vspace{-10pt}
\end{wraptable}

Beyond editing, temporal and robustness benchmarks such as TiEBe \citep{almeida2025tiebe} and RECALL \citep{liu2023recall}, as well as diagnostic frameworks such as KScope \citep{xiao2025kscope}, evaluate evolving or externally provided knowledge, but do not provide controlled evaluation of how updates are incorporated and used across tasks. A complementary line of work (e.g., Bloom-style benchmarks \citep{huber2025bloom1, chen2026bloom2} and MentorQA \citep{bhalerao2026mentorqa}) assesses general reasoning capabilities across tasks, but does not isolate individual factual updates or their downstream effects. As a result, existing benchmarks primarily focus on factual recall, with limited evaluation of generalization beyond isolated settings or single task formats (Table~\ref{tab:editing_benchmarks}).

In contrast, \textbf{MedKIT} targets \emph{factual knowledge integration}. We operationalize a multi-level generalization framework through fact-centered evaluation probes that systematically assess how newly integrated knowledge transfers across lexical, relational, compositional, and operational settings. This enables a shift from evaluating whether models remember updated facts to assessing whether they can \emph{use} them consistently across clinically motivated tasks.

\textbf{Knowledge Editing.}
Early knowledge editing approaches such as ROME \citep{meng2023rome} and MEMIT \citep{Meng2022memit} modify localized representations to update specific facts, while more recent variants (e.g., AlphaEdit \citep{fang2025alphaedit}, WISE \citep{Wang2024wise}, MedLaSA \citep{xu2024medlasa}) improve robustness and reduce interference across edits. Alternative approaches include learned editors (MEND \citep{Mitchell2021mend}), memory-based methods (SERAC \citep{mitchell2022serac}, GRACE \citep{Hartvigsen2022grace}, MEMOIR \citep{wang2026memoir}), and in-context editing techniques (IKE \citep{zheng2023ike}). These methods provide fine-grained control over individual updates, but are typically evaluated on recall-centric benchmarks.

\textbf{Continual Post-Training.}
Factual updates can also be integrated through continued post-training on new data. This includes supervised fine-tuning (SFT), often implemented with parameter-efficient adapters (e.g., LoRA \citep{hu2021lora}, O-LoRA \citep{wang2023olora}, SEEKR \citep{he2024seekr}) or preference-based optimization methods such as DPO \citep{rafailov2023direct} and GRPO \citep{deepseek-math}. Model merging provides an additional mechanism for combining updated and base models while mitigating drift \citep{wortsman2022robustfinetuningzeroshotmodels, ilharco2022patching, ramé2024warm}. These approaches scale naturally to batches of updates, but operate at the level of datasets or behaviors rather than individual facts, limiting control over how specific updates propagate and generalize.

\textbf{Retrieval Augmented Generation (RAG).} 
Retrieval-augmented approaches incorporate new knowledge by storing it externally and retrieving it at inference time \citep{lewis2020rag1, guu2020realm, karpukhin2020dense}. This enables dynamic updates without modifying model parameters and achieves strong performance on knowledge-intensive tasks \citep{borgeaud2022improving, izacard2022atlas}. Recent work extends this paradigm to agentic behavior, in which models iteratively plan search actions and refine retrieval through interaction (e.g., ReAct \citep{yao2023react}). While these approaches improve access to relevant information, they rely on effective retrieval and do not enforce consistent integration of knowledge into the model’s internal representations.

\section{MedKIT Benchmark Construction and Evaluation}

In this section, we describe the MedKIT benchmark, including its data source, preprocessing pipeline, and task design for evaluating the \textbf{generalization of knowledge integration} under realistic clinical settings with evolving evidence. Figure~\ref{fig:benchmark_construction} summarizes the construction and composition of MedKIT. The benchmark contains 6,196 updates spanning publications from 1960 to 2026, covering 2,135 treatment regimens across 329 clinical conditions and 12 oncology groups, with a balanced label distribution (28\% \textit{superior}, 43\% \textit{no difference}, 29\% \textit{inferior}). Each update is associated with a publication timestamp, enabling temporally ordered update sequences at different granularities (e.g., daily or weekly). Figure~\ref{fig:benchmark_overview} presents additional benchmark statistics, with further details reported in Appendix~\ref{app:extended_stats}. 

\subsection{Data Source}
MedKIT is constructed from \textit{HemOnc.org} \citep{warner2019hemonc}, a curated oncology knowledge base of evidence-backed treatment comparisons with associated publication dates.
HemOnc.org represents clinical knowledge as pairwise comparisons between treatment options within specific clinical settings, reflecting the comparative structure of clinical evidence.
Each entry specifies a condition, clinical context, two treatments, and their relative efficacy with respect to a given endpoint. For example, a record may indicate that ``\textit{Fluorouracil + Leucovorin (FULV) is inferior to Irinotecan + Fluorouracil + Leucovorin (IFL) for colon cancer in the adjuvant setting in terms of disease-free survival (DFS).}''
This representation directly maps to \emph{fact-centered update instances}, in which each comparison corresponds to a single piece of evidence with a precise publication timestamp.
It thus provides a natural foundation for evaluating how models integrate new knowledge and apply it beyond direct recall.

\subsection{Benchmark Construction Pipeline}
We transform structured clinical comparisons of HemOnc.org into \emph{factual update instances}, each paired with diverse evaluation probes, via a four-stage construction pipeline.

\myparagraph{Step 1: Extraction and Structuring.}
We extract pairwise treatment comparisons describing the relative efficacy of two regimens for a given condition, clinical context, and endpoint. Each instance is represented as a structured tuple of the form \textit{(Treatment A, relation, Treatment B, Condition, Context, Endpoint)}, corresponding to a single fact-centered knowledge update derived from a specific clinical study. We filter out incomplete or ambiguous entries and deduplicate equivalent comparisons to retain only well-defined updates.

\myparagraph{Step 2: Canonicalization.}
All tuples are normalized into a consistent representation by standardizing entities and mapping relations to a fixed label space (superior, inferior, no difference). When multiple endpoints are reported, we select a single canonical instance using a fixed priority scheme based on endpoint type and clinical relevance. Applied to the full HemOnc.org snapshot, this yields 6,196 canonical updates, comprising the complete set of eligible comparisons after preprocessing rather than a sampled subset.

\myparagraph{Step 3: Evidence Augmentation.}
Each update is augmented with its supporting clinical evidence by retrieving the corresponding PubMed abstracts. This provides an evidence-grounded representation that supports both parametric integration methods and retrieval-based approaches.

\myparagraph{Step 4: Task Generation.}
From each update instance, we construct a set of evaluation tasks. All tasks are derived from the same underlying fact but vary in their formulations and output spaces, spanning both closed-form classification and open-ended generation. This design enables systematic evaluation of how integrated knowledge transfers across task formats and levels of generalization.

\begin{figure*}[t]
    \centering
    \includegraphics[width=\textwidth]{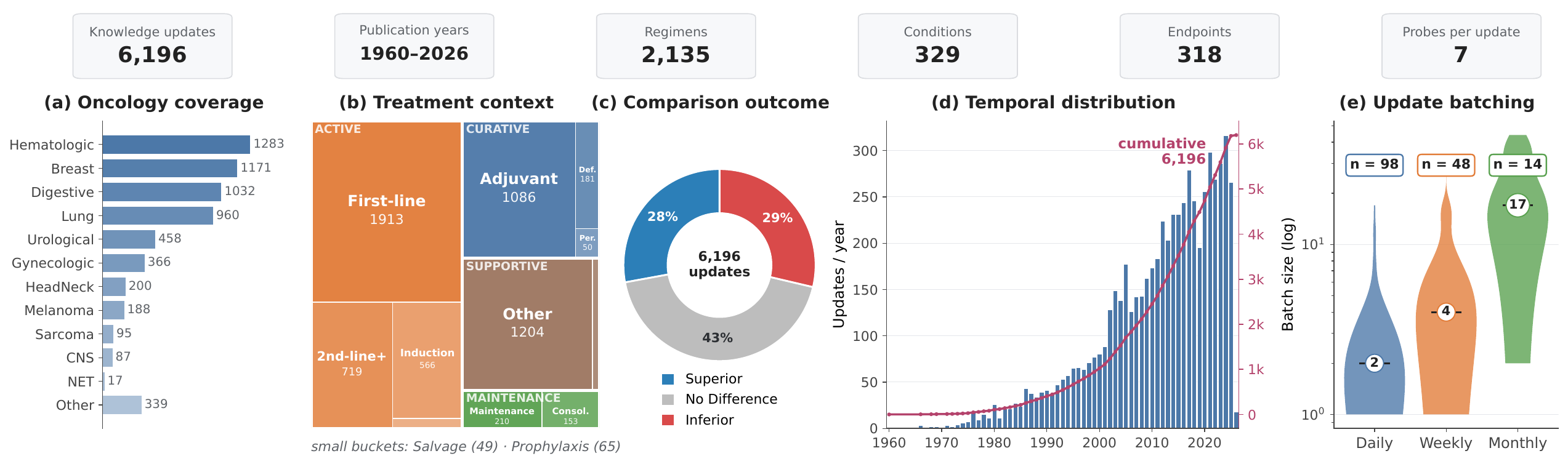}
    \caption{
    \textbf{MedKIT statistics and temporal structure.} The benchmark spans 6,196 clinically grounded factual updates across conditions and oncology groups, with balanced labels and temporally ordered batches that enable realistic sequential evaluation.  
    }
    \label{fig:benchmark_overview}
    \vspace{-10pt}
\end{figure*}

\subsection{Task Design}
Our task design follows a \textbf{fact-centered evaluation protocol}, in which each update instance is evaluated through a set of targeted probes derived from the same underlying clinical comparison. This stands in contrast to standard train/test splits, which assess generalization across disjoint samples rather than at the level of individual knowledge updates.
All task variants are generated from a fixed set of templates manually designed in collaboration with an MD clinician to ensure clinically plausible and semantically consistent task formulations. This design isolates specific forms of generalization while avoiding confounding variation introduced by LLM-generated prompts.
We provide representative examples in Figure~\ref{fig:benchmark_construction} and full templates in Appendix~\ref{app:task_construction}.

\myparagraph{Update Task.}
Each update instance is anchored by a canonical \textit{closed-form comparison task}, formulated as a three-way classification problem. 
This task directly probes whether the updated fact has been successfully integrated and serves as the primary measure of update success, analogous to a training-set accuracy for the integrated update.

\myparagraph{Generalization Tasks.}
To assess how integrated knowledge transfers beyond the update task, we construct four complementary task variants for each update instance, capturing progressively more demanding forms of generalization, building on prior work on the multidimensional nature of generalization in NLP \citep{hupkes2023taxonomy} (see Fig.~\ref{fig:benchmark_construction} or App.~\ref{app:task_construction} for examples).

\textit{Lexical} generalization tests robustness to surface-level variation by using paraphrased versions of the update question with identical semantics. 
\textit{Relational} generalization tests whether the model has internalized the relationship between entities by reversing the treatment order in the comparison. 
\textit{Compositional} generalization tests whether the updated fact can be used within a reasoning process through open-form comparison questions requiring a factually correct response. 
\textit{Operational} generalization tests whether the updated knowledge is 
applied implicitly in open-ended treatment recommendations, without explicitly prompting for the comparison.

\myparagraph{Locality.}
Here, the aim is to evaluate whether integrating a specific update affects unrelated knowledge. For each update instance, we construct a probe from the same oncology group but with a different condition and clinical context, yielding a \emph{similar yet factually independent} sample. This allows us to assess whether models preserve nearby, unrelated knowledge under targeted updates.

\subsection{Evaluation Protocol}
We evaluate model behavior across all task types using a combination of exact-match metrics and LLM-based judges, depending on the structure of the task outputs.

\textbf{Closed-form evaluation.}
For lexical and relational tasks where the answer could be \textit{superior}, \textit{inferior}, or \textit{no difference}, we use exact match accuracy. If the model output does not explicitly match one of these labels (e.g., due to free-form phrasing), we fall back to an LLM-based judge to determine whether the response correctly implies the ground-truth answer.

\textbf{Open-form evaluation with LLM judges.}
For compositional, operational, and locality tasks, we use an LLM-based judge to assess consistency with the ground-truth comparison. For \textit{compositional and locality tasks}, the judge evaluates whether the model expresses the correct treatment preference (or equivalence), scoring responses on a 5-point Likert scale from misaligned to fully aligned. For \textit{operational tasks}, the judge evaluates whether the generated recommendation is consistent with the underlying comparison, thereby testing the use of implicit knowledge in open-ended settings.
Judge prompts were iteratively refined with a medical expert to ensure a domain-appropriate and consistent evaluation (see full prompt details in Appendix~\ref{app:judge_evaluation}).

\myparagraph{Judge validation.}
\label{subsec:judge_validation}
Because our open-form evaluation relies on LLM judgments, we validate its robustness and clinical grounding through three complementary studies (App.~\ref{app:judge_validation}).

\emph{Study A: Cross-judge robustness.} We first test whether comparative conclusions are robust to the choice of judge. Because many methods cluster near pre-update performance on the open-form tasks, we compare performance tiers rather than fine-grained rankings. All seven judges recover the three-tier structure on the compositional task, and five of seven recover the oracle-vs-rest separation on the operational task. The remaining two show only one boundary-level disagreement between statistically indistinguishable methods. Independently derived tiers likewise show substantial agreement (Fleiss' $\kappa=0.61$ and $0.70$).

\emph{Study B: Clinician agreement.} Two independent clinicians score the same 100 held-out responses using the judge rubric. The clinicians show clear agreement on compositional and operational tasks (Spearman's $\rho=0.76$ and $0.62$), while \texttt{gpt-4o} closely agrees with their consensus ($\rho=0.83$ and $0.64$). Judge deviations concentrate on examples where the clinicians themselves disagree: when both clinicians agree, \texttt{gpt-4o}'s mean absolute deviation from their consensus is 0.31 and 0.43, compared with 1.82 and 1.32 when they disagree.

\emph{Study C: Intra-judge stability.} Finally, we query each judge five times on a deterministic subset of examples. Repeated scoring is highly stable, with median per-item variance equal to zero for nearly all settings. Additional analyses find no evidence of systematic verbosity or generator-family bias.

Together, these studies show robust comparative conclusions, agreement with independent clinicians, and stable repeated evaluation. We select \texttt{gpt-4o} as the primary judge based on its overall trade-off between clinical grounding, comparative robustness, and inference cost.

\section{Experiments}

We evaluate a diverse set of 12 approaches spanning three families: knowledge editing, including parameter-editing methods (AlphaEdit, MEMIT) and augmented editing methods (WISE, GRACE, MEMOIR, IKE), continual post-training (LoRA-Merge, O-LoRA, SEEKR), and retrieval-based methods (BM25-RAG, Dense-RAG, Agentic-RAG). Detailed descriptions of all methods are provided in Appendix~\ref{app:method_overview}.
Experiments are conducted on five instruction-tuned LLMs across two scales (4B and 8B), including both general-purpose (Gemma-3 \citep{gemmateam2025gemma3technicalreport}, Qwen-3 \citep{qwen3technicalreport}, LLaMA-3.1 \citep{grattafiori2024llama3herdmodels}) and medical-adapted variants (MedGemma \citep{medgemma2024}, Bio-Medical-LLaMA-3 \citep{ContactDoctor_Bio-Medical-Llama-3-8B}). We create 48 weekly update batches comprising 283 updates from 2025 onward, minimizing the likelihood that these facts were seen during pretraining.

For each update, we evaluate all task variants and locality probes, reporting post--pre performance differences in percentage points (pp) relative to the base model, with pre-update performances reported in Appendix~\ref{app:pre_edit_analysis}. LLM-judge scores are normalized to $[0,1]$, and hyperparameters are tuned per (method × model) on held-out updates. Implementation details are provided in Appendix~\ref{app:compute_implementation}, with complete numerical results in Appendix~\ref{app:complete_results}.

\subsection{Generalization Beyond Recall}
\label{subsec:generalization}

\begin{figure*}[t]
    \centering
    \includegraphics[width=\textwidth]{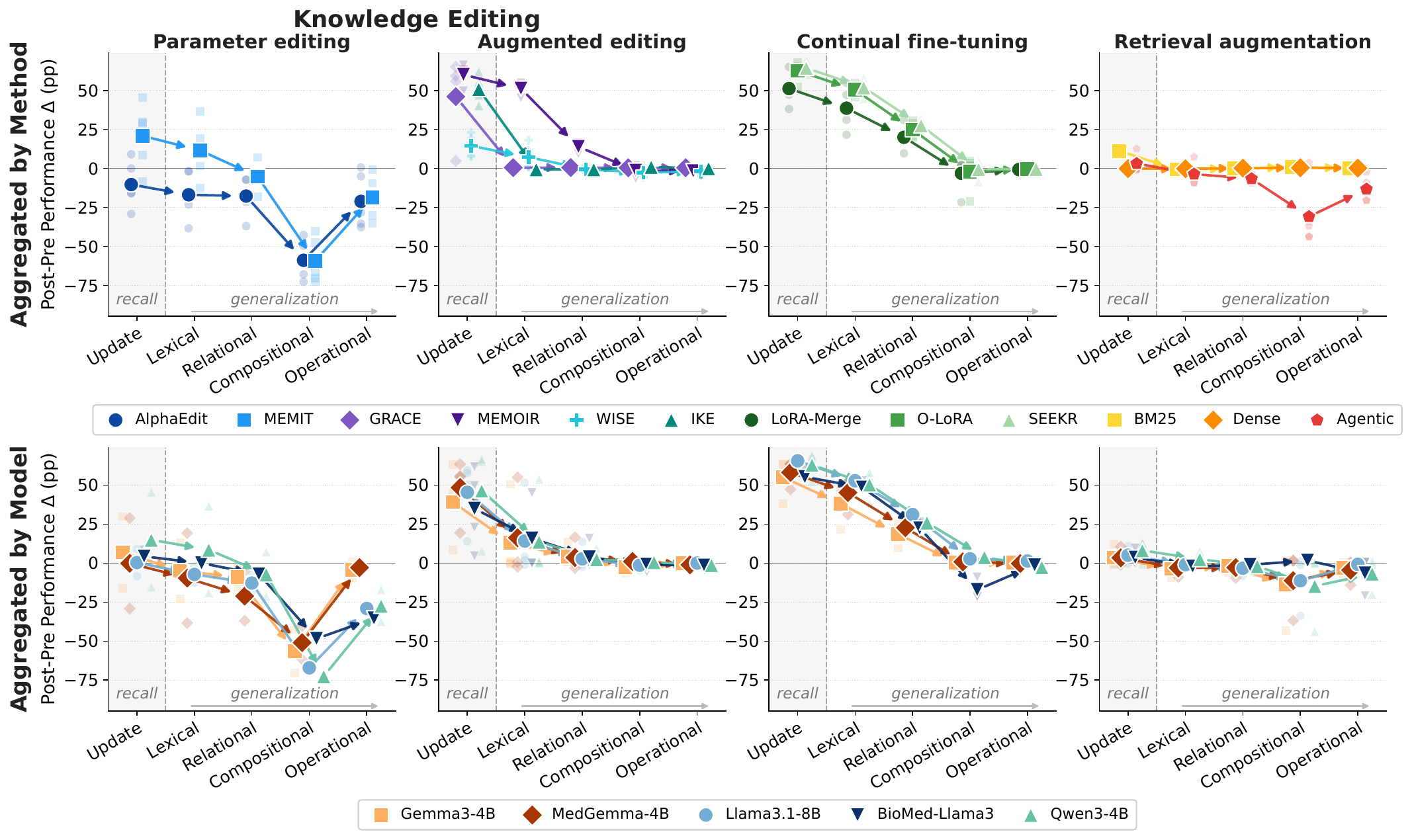}
    \caption{
    \textbf{Generalization of integrated knowledge across task formats.}
    Mean post--pre performance ($\Delta$) across tasks from direct recall (\textit{Update}) to increasingly demanding generalization settings. Top: aggregated by method; bottom: aggregated by model, with faint markers indicating individual model--method combinations.
    Across both views, performance consistently degrades beyond lexical variation, indicating that improvements in recall do not translate into reliable generalization across reformulated and open-ended task settings.
    }
    \label{fig:generalization_combined}
    \vspace{-10pt}
\end{figure*}

Figure~\ref{fig:generalization_combined} shows mean post--pre performance across the generalization hierarchy, from direct recall to increasingly demanding downstream tasks.

\myparagraph{Integrated knowledge shows limited generalization beyond surface form.}
Aggregated by method (top row), all approaches exhibit a consistent degradation from the update task to more demanding generalization settings, but the shape of this decay differs fundamentally across method families. 
Importantly, this degradation is not simply due to task difficulty. 
With the correct evidence provided in context, models achieve substantial gains across all task types (e.g., +20pp to +50pp), indicating that the observed failures primarily stem from limited generalization rather than inherent task complexity (see App.~\ref{app:rag_ablations}).

While some knowledge editing methods achieve strong gains on the update task, generalization beyond the original formulation remains limited across the family.
Parameter-editing approaches (MEMIT, AlphaEdit) prove to be unstable: MEMIT achieves only modest gains on the update task (+21pp), while AlphaEdit is inconsistent (-9pp), with both exhibiting large negative shifts on compositional tasks (down to -59pp). 
This pattern reflects known instability in parameter-editing methods, where sequential updates accumulate interference and degrade downstream behavior \citep{gupta2024scale, Yang2024butterfly, thede2025wikibigedit}.

Augmented editing methods, relying on auxiliary memory or routing, show improved stability. However, this stability does not translate into improved generalization.
Lookup-based methods such as GRACE and IKE achieve strong gains (+46pp and +51pp) on the update task but drop to near-baseline under reformulation (+0.5pp and -0.6pp). This pattern is consistent with the lookup mechanism being triggered only by the original update, or failure to retrieve the relevant stored update. WISE offers slightly improved robustness to lexical variation (+7pp) through routing, but still collapses at the relational level (-1pp).
MEMOIR is the strongest augmented-editing baseline, achieving the largest update gains (+60pp) and maintaining high lexical robustness (+51pp). However, it still fails to generalize further: performance drops at the relational level (+14pp) and remains near baseline on compositional (-1pp) and operational (-2pp) tasks, indicating that improved recall does not translate into recomposition in downstream reasoning.

Continual post-training exhibits a smooth, monotonic decay from update to operational tasks and achieves the strongest overall generalization performance. It yields mean improvements of +49pp on the update task, +36pp on lexical variants, and +19pp at the relational level. This pattern indicates that a more holistic, global integration of updates, rather than local parameter editing, favors generalization. However, these gains do not extend to the more demanding compositional and operational tasks, where performance remains close to baseline. We additionally evaluate preference optimization (DPO, GRPO), which proves less effective for knowledge integration, following a similar but consistently weaker trend (see App.~\ref{app:pref_rl_baselines}).

Retrieval-based approaches provide only limited improvements on the update task (+11pp for BM25) and remain flat across all downstream tasks. We identify two main failure modes: (i) insufficient retrieval 
accuracy (e.g., recall@3 $\approx$ 40\% for BM25), and (ii) limited ability of the model to incorporate retrieved evidence into reasoning (see Appendix~\ref{app:rag_ablations}). Notably, agentic RAG shows similar behavior, indicating that this agentic retrieval variant does not substantially improve generalization at the given model scale.

\begin{wrapfigure}{r}{0.45\textwidth}
    \centering
    \includegraphics[width=\linewidth]{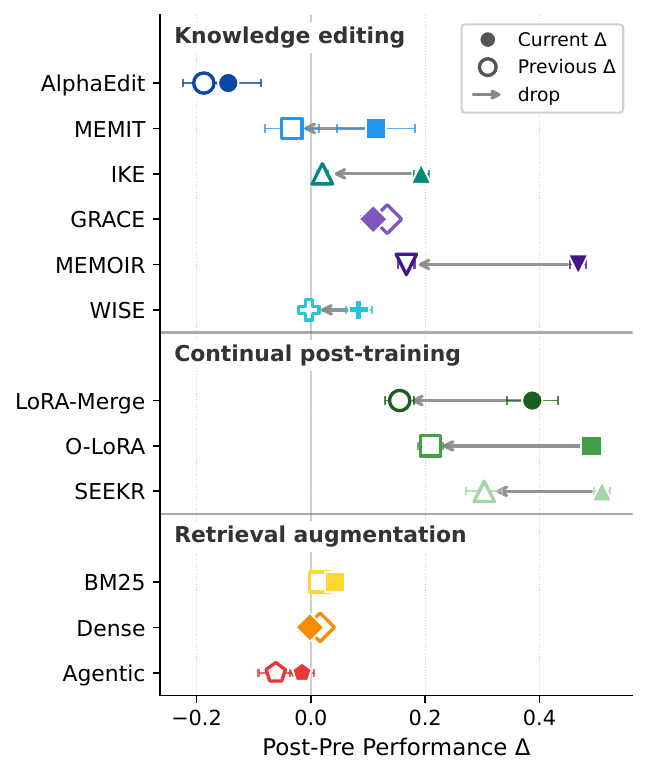}
    \vspace{-10pt}
    \caption{
    Retention under sequential updates. Current $\Delta$ measures immediate gains, Previous $\Delta$ performance on earlier updates.
    Retention varies by method, reflecting differences in underlying integration mechanisms.
    }
    \label{fig:past_performance}
    \vspace{-15pt}
\end{wrapfigure}

\myparagraph{Generalization limitations are consistent across models.}
Aggregated by model (bottom row), we observe highly consistent generalization behavior across all model families. Independent of architecture or domain specialization, all models exhibit the same characteristic degradation from the update task to downstream settings: strong gains on the update task, partial transfer to lexical and relational variants, and near-baseline or negative performance on compositional and operational tasks. This consistency indicates that the observed generalization behavior is primarily determined by the integration method rather than the underlying model.

We further find no structural differences between general-purpose and medical-adapted models. For example, MedGemma closely mirrors Gemma-3 across all task levels, while Bio-Medical-LLaMA-3 follows nearly the same trajectory as LLaMA-3.1. Despite their domain specialization, both models exhibit the same degradation patterns beyond lexical variation and similarly limited gains on compositional and operational tasks.
Only minor quantitative differences emerge. Gemma-based models exhibit somewhat smaller operational degradation for parameter-editing methods (e.g., around -3pp to -4pp compared to up to -35pp for Qwen-3), indicating slightly more stable downstream behavior. However, these differences do not alter the overall generalization patterns.

\myparagraph{Generalization patterns remain stable across temporal update regimes.}
We additionally analyze the impact of batching strategies. Comparing weekly (48 batches) and daily (98 batches) updates, we find that the generalization profiles remain stable across all tested regimes. While for some methods absolute performance varies slightly with batch size, the relative ranking and qualitative behavior of methods are unchanged (see Appendix~\ref{app:batching_ablations} for details).

 \begin{tcolorbox}[
  colback=white,
  colframe=takeawaysColor,
  boxrule=1.5pt,
  arc=3pt,
  left=4pt,right=4pt,top=4pt,bottom=4pt,
]
\textbf{\textcolor{takeawaysColor}{Takeaways.}}
Updates improve recall, but this rarely translates into usable knowledge. Continual post-training shows the strongest transfer beyond lexical variation; knowledge editing either fails under reformulation or causes interference, and retrieval is limited by recall and utilization. Strong recall does not imply usable knowledge.
\end{tcolorbox}

\subsection{Retention Under Sequential Updates}
We test the model after every second batch on a sentinel pool of earlier samples, comparing immediate post-edit performance with retained performance (details in App.~\ref{app:reporting_metric}). 

\myparagraph{Retention of past updates varies across methods and mechanisms.}
Figure~\ref{fig:past_performance} shows that retention behavior differs substantially across methods, rather than following a uniform trade-off between immediate gains and stability. 
Knowledge editing methods are highly heterogeneous. MEMOIR achieves strong post-edit performance (+48pp) but suffers the largest degradation over time (drop of 32pp), indicating interference within its memory mechanism. In contrast, GRACE maintains nearly identical post- and past-performance (+11pp vs. +13pp), demonstrating stable retention through localized updates, albeit at the cost of smaller initial gains.

Continual post-training achieves the largest immediate improvements (up to +50pp) but exhibits substantial degradation on past samples (e.g., -20pp to -30pp), consistent with catastrophic forgetting under sequential updates. To mitigate this effect, these methods incorporate mechanisms such as weight merging or replay of past samples. Among them, SEEKR retains the most (past $\Delta$ +30pp), reflecting the benefit of its explicit replay-based memory. However, in the setting of factual knowledge integration, where updates correspond to largely independent facts, replay remains inherently limited: reinforcing one fact does not support retention of others.
Retrieval-based methods remain effectively unchanged over time, with both post- and past-performance close to baseline (within $\pm$4pp), as they do not modify model weights and therefore exhibit minimal interference across updates.

 \begin{tcolorbox}[
  colback=white,
  colframe=takeawaysColor,
  boxrule=1.5pt,
  arc=3pt,
  left=4pt,right=4pt,top=4pt,bottom=4pt,
]
\textbf{\textcolor{takeawaysColor}{Takeaways.}}
Retention depends on the underlying integration mechanism rather than a fixed trade-off. Continual post-training exhibits retention losses consistent with inter-batch interference, augmented editing preserves updates with limited integration, and retrieval-based methods remain stable without parameter updates. No method achieves both strong integration and stable retention.
\end{tcolorbox}

\begin{figure*}[t]
    \centering
    \begin{subfigure}[t]{0.33\textwidth}
        \centering
        \includegraphics[width=\linewidth]{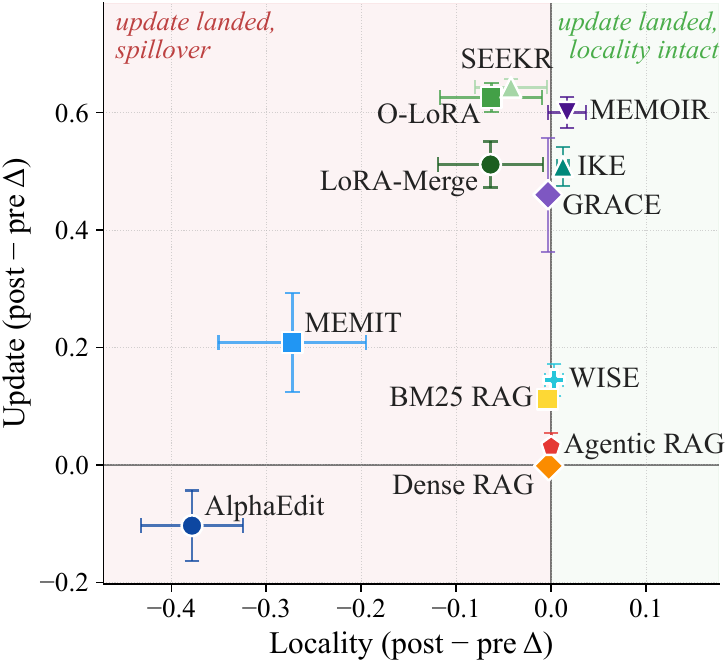}
        \caption{Within-domain locality.}
        \label{fig:locality_spillover}
    \end{subfigure}%
    \hfill
    \begin{subfigure}[t]{0.66\textwidth}
        \centering
        \includegraphics[width=\linewidth]{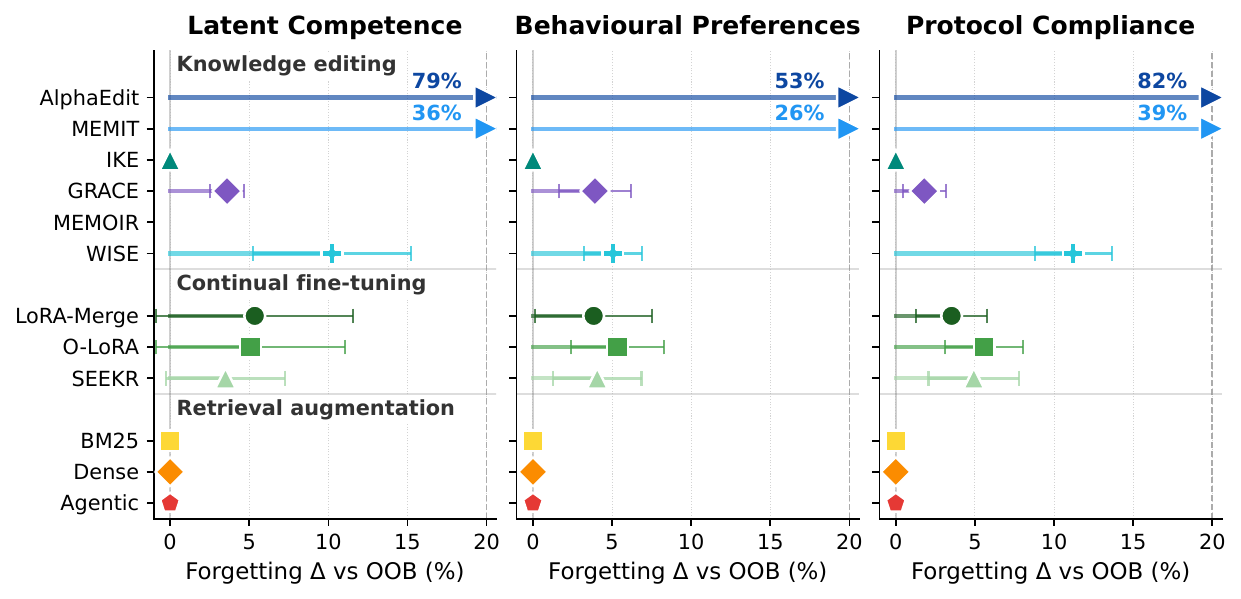}
        \caption{General capability drift.}
        \label{fig:captrack_placeholder}
    \end{subfigure}

    \caption{
    \textbf{Capability preservation under knowledge integration.}
    Left: Update success versus locality change, measuring whether methods integrate target updates while preserving neighboring oncology facts. 
    Right: Changes in general capabilities, grouped into latent competence, behavioral preferences, and protocol compliance, highlight the risk of updates interfering with the model.
    }
    \label{fig:capability_preservation}
    \vspace{-10pt}
\end{figure*}

\subsection{Knowledge and Capability Preservation}
\label{subsec:experiments_preservation}
We evaluate whether integrating new knowledge affects unrelated facts and general capabilities. MedKIT provides locality probes to measure in-domain spillover. We further use CapTrack \citep{thede2026captrack} to assess changes on a broad set of general capability probes. Figure~\ref{fig:capability_preservation} reports locality and capability-level forgetting, averaged across models.

\myparagraph{Locality violations are concentrated in parameter-editing methods.}
Figure~\ref{fig:capability_preservation}a reveals clear differences in how methods affect neighboring knowledge. Parameter-editing methods (MEMIT, AlphaEdit) exhibit substantial locality violations, with large negative shifts indicating strong spillover to related facts, consistent with the instability observed in Sec.~\ref{subsec:generalization}. 
In contrast, augmented editing methods (GRACE, MEMOIR, WISE) largely preserve locality, with near-zero or slightly positive changes, indicating better preservation of neighboring facts. Continual post-training shows mild degradation, reflecting limited but non-negligible spillover. Retrieval-based methods remain stable, as they do not modify model parameters and therefore exhibit minimal interference.

\myparagraph{Parameter updates risk degrading general capabilities.}
These effects extend beyond the medical domain to broader language-model capabilities (Figure~\ref{fig:capability_preservation}b). Consistent with their strong locality violations, parameter-editing methods exhibit substantial degradation across latent competence, behavioral preferences, and protocol compliance. Augmented editing methods are generally more stable, though WISE still shows notable degradation in competence ($-10\%$) and protocol compliance ($-11\%$). Continual post-training methods occupy an intermediate regime, preserving much of the underlying competence while introducing moderate behavioral and execution drift ($-3.5\%$ to $-5.6\%$). 
In contrast, retrieval-based methods and IKE leave model parameters unchanged and therefore exhibit negligible degradation across capability dimensions. Additional analysis of capability preservation as well as training and inference costs is provided in Appendix~\ref{app:capability_preservation} and Appendix~\ref{app:deployment_cost}.

 \begin{tcolorbox}[
  colback=white,
  colframe=takeawaysColor,
  boxrule=1.5pt,
  arc=3pt,
  left=4pt,right=4pt,top=4pt,bottom=4pt,
]
\textbf{\textcolor{takeawaysColor}{Takeaways.}}
Modifying model parameters risks both locality violations and capability degradation. Parameter-editing methods cause strong spillover and broad capability loss, while continual post-training introduces milder but systematic shifts. Methods that avoid parameter updates remain stable, highlighting the challenge of strong integration without compromising the model.
\end{tcolorbox}

\subsection{Generalization Beyond MedKIT}
\label{sec:cross_domain}

\begin{wraptable}{r}{0.50\textwidth}
\vspace{-8pt}
\centering
\small
\caption{Cross-domain evaluation on WikiBigEdit. Mean post--pre change in exact-match containment (pp) across two models.}
\label{tab:wikibigedit_main}
\setlength{\tabcolsep}{4pt}
\begin{tabular}{lrrrr}
    \toprule
    Method & Update & Lexical & Compos. & Locality \\
    \midrule
    GRACE  & +75.5 & +0.0  & +0.0 & +0.0 \\
    MEMIT  & +38.1 & +22.8 &  0.0 & -1.9 \\
    MEMOIR & +52.2 & +38.1 & -0.8 & -2.2 \\
    SEEKR  & +41.9 & +20.2 & -1.9 & -0.9 \\
    \bottomrule
\end{tabular}
\vspace{-6pt}
\end{wraptable}

Our results reveal a substantial gap between integrating new knowledge and using it in more demanding settings. To test whether this gap is specific to MedKIT's clinical domain and pairwise formulation, we additionally evaluate four representative methods on WikiBigEdit~\citep{thede2025wikibigedit}, a benchmark of evolving Wikidata facts, using LLaMA-3.1-8B and Qwen-3-4B. We map direct recall, paraphrases, multi-hop reasoning, and neighborhood preservation to our \emph{update}, \emph{lexical}, \emph{compositional}, and \emph{locality} dimensions. 

Table~\ref{tab:wikibigedit_main} shows the same qualitative pattern as MedKIT. All methods substantially improve direct recall ($+38$--$75$ pp), and several transfer these gains to lexical variations. However, these improvements largely disappear on compositional reasoning, despite locality remaining near pre-update performance.

\begin{tcolorbox}[
  colback=white,
  colframe=takeawaysColor,
  boxrule=1.5pt,
  arc=3pt,
  left=4pt,right=4pt,top=4pt,bottom=4pt,
]
\textbf{\textcolor{takeawaysColor}{Takeaways.}}
The gap between factual recall and knowledge utilization extends beyond MedKIT. On WikiBigEdit, methods successfully integrate updated facts and often generalize to lexical variations, yet these gains largely disappear when the same knowledge must be used compositionally.
\end{tcolorbox}

\section{Conclusion}

We introduce MedKIT, a benchmark for evaluating knowledge integration in large language models under realistic sequences of evolving clinical evidence. By combining factual updates with a multi-level generalization framework, MedKIT enables a fine-grained analysis of how newly integrated knowledge is used beyond direct recall.

Our results reveal a consistent gap between \emph{integrating} knowledge and \emph{using} it. While many methods achieve strong gains on the original update task, these improvements rarely generalize reliably beyond the original formulation. Sequential updates further expose limitations of current approaches: methods that integrate updates effectively often struggle to retain past knowledge or preserve unrelated knowledge and broader model capabilities.

We position MedKIT as a testbed for developing methods that support structured generalization, enabling models to reliably apply newly integrated knowledge across tasks and contexts while preserving existing capabilities. By grounding updates in real-world clinical evidence and temporally evolving information, MedKIT provides a realistic setting for studying how language models can be maintained under changing knowledge.

MedKIT is intended for method development rather than clinical deployment. It captures selected aspects of evolving clinical evidence and should not be used to support clinical decision-making or treatment recommendations in practice. The benchmark also inherits biases from the underlying clinical literature and source data. Additional discussion of limitations, broader considerations, and deployment implications is provided in Appendix~\ref{app:limitations_ethics}.

\newpage

\section*{Acknowledgements}

Lukas Thede thanks the International Max Planck Research School for Intelligent Systems (IMPRS-IS) for support. We are grateful for support by the Carl Zeiss Foundation, project "Certification and Foundations of Safe Machine Learning Systems in Healthcare". This work was partially funded by the ERC (853489 - DEXIM) and the Alfried Krupp von Bohlen und Halbach Foundation, which we thank for their generous support. We thank clinicians Elise Thede and Maximilian N\"agele for contributing their medical expertise and for their support in the validation of MedKIT. The authors declare no competing interests.

\bibliography{references}
\bibliographystyle{plainnat}


\newpage
\appendix

\section{Benchmark Construction}
\label{app:benchmark_construction}

This appendix summarizes the construction of MedKIT. The released code contains the full implementation details; here, we focus on the main preprocessing decisions needed to interpret the benchmark.

\subsection{Data Source}
\label{app:data_sources}
MedKIT is built from a fixed HemOnc.org \citep{warner2019hemonc} knowledge-base snapshot (2026-03-12). HemOnc.org provides clinically curated comparisons of oncology treatments grounded in published evidence. Each benchmark instance is derived from a structured comparison between two treatment regimens for a given condition, clinical context, and endpoint.

For each instance, we store the treatment pair, condition, context, endpoint, standardized outcome label, publication date, and supporting evidence. Outcome labels are mapped to the three-way label space
\[
\{\textit{superior}, \textit{inferior}, \textit{no difference}\}.
\]
Publication dates are obtained from the associated study records and define the temporal order used in the sequential update setting. Supporting evidence consists of linked PubMed titles and abstracts; rows without resolvable evidence are discarded.

\subsection{Preprocessing and Canonicalization}
\label{app:preprocessing}

We apply an entirely rule-based preprocessing pipeline; no learned or generative model is used at any stage.
We remove incomplete rows, self-comparisons, uninformative efficacy statements, and entries without supporting evidence. We then normalize efficacy descriptions to the three target labels using curated mappings and rule-based fallbacks. Ambiguous labels that cannot be mapped reliably are excluded.

When a study reports multiple endpoints for the same treatment comparison, we retain a single canonical endpoint. We prioritize endpoint type
\[
\text{Primary} > \text{Co-primary} > \text{Secondary} > \text{Undesignated}
\]
and then clinically relevant endpoints, with OS and PFS ranked highest. In rare symmetric comparisons where either treatment ordering can serve as the canonical representation, we use seed~$42$ to resolve the ordering reproducibly. This yields one atomic update per study-level comparison.

We use curated treatment names from HemOnc directly, augment conditions with disease stage where available, and assign each condition to an oncology group for locality construction. For every canonical comparison, we also construct the reverse direction by swapping treatments and flipping the label. Exact duplicates are removed, and tuples with inconsistent labels within the same evidence setting are discarded.

\subsection{Conflict Handling}
\label{app:conflict_handling}

Some clinical comparisons appear multiple times across studies or publication dates with different outcomes. We flag such cases as \emph{conflicting} when the same condition, context, endpoint, regimen, and comparator occur with more than one label.

These conflicts reflect genuine changes or disagreements in clinical evidence rather than data errors. We retain them in the released benchmark with a \texttt{conflicting\_edit} flag, but filter them from the main experiments. This keeps the headline evaluation focused on unambiguous updates while preserving contested cases for future analysis.

\subsection{Task Construction}
\label{app:task_construction}
Each MedKIT instance is associated with a set of task variants derived from the same underlying clinical comparison, enabling controlled evaluation of knowledge integration across different query formulations.

\paragraph{Fact-centered evaluation.}
All task variants for a given instance are derived from the same underlying clinical comparison. This design isolates the effect of task formulation, allowing us to evaluate whether newly integrated knowledge generalizes across different query types, rather than across disjoint train--test splits.

\paragraph{Generalization tasks.}
We construct five task types corresponding to the generalization hierarchy introduced in the main paper:
\begin{itemize}
    \item \textbf{Update (closed QA):} the canonical comparison task with a fixed three-way answer space (\textit{superior}, \textit{inferior}, \textit{no difference}).
    \item \textbf{Lexical:} two paraphrased variants of the anchor question that preserve semantics while varying surface form.
    \item \textbf{Relational (mirror):} the comparison with treatments swapped and the label flipped, testing whether the model has internalized the relationship.
    \item \textbf{Compositional (open QA):} a structured open-ended comparison task in which the model must reason about two treatments given a condition, context, and endpoint.
    \item \textbf{Operational (open Gen.):} an open-ended generation task in which the model must produce a context-appropriate treatment recommendation without being explicitly prompted to compare the treatments.
\end{itemize}

All generalization tasks are generated deterministically from the same structured tuple, using templates manually designed in collaboration with an MD clinician, ensuring that differences in performance reflect changes in task formulation rather than in the underlying data.

\paragraph{Locality task.}
In addition to generalization, each update is paired with a \emph{locality} task that tests whether integrating the update affects unrelated knowledge. The locality instance is a structurally similar comparison drawn from the same distribution but with a different condition (and typically different context and label). Preference is given to examples within the same oncology group; if none are available, constraints are relaxed while maintaining factual independence.

\paragraph{Prompt Templates.}
Listing~\ref{lst:task_examples} shows the exact prompts used in evaluation for a representative instance. Closed-form tasks share a common instruction block with a constrained answer space, while open-form tasks use separate instructions for reasoning and generation.

\begin{lstlisting}[caption={Full prompt templates for a representative MedKIT instance.},
label=lst:task_examples, basicstyle=\ttfamily\small, breaklines=true,
columns=fullflexible, frame=single]

=== Closed-form tasks (Update / Lexical / Relational) ===

You are a knowledgeable medical assistant supporting oncologists and hematologists in evaluating treatment options.

Given a clinical question, compare two treatment options with respect to a specific outcome endpoint and determine their relative efficacy based on current clinical evidence.

Answer with exactly one of the following three options:
- superior      (Treatment 1 outperforms Treatment 2 on the specified endpoint)
- inferior      (Treatment 1 underperforms Treatment 2 on the specified endpoint)
- no difference (no meaningful difference between the two treatments on the specified endpoint)

Do not include any explanation or additional text.

[Update]
Task:
Choose an option that best describes the PFS outcome of Observation compared to ITd when used to treat Multiple myeloma (Non-curative second-line consolidation therapy).
Response:

[Lexical (paraphrase)]
Task:
Select the option that most accurately reflects the PFS outcome of Observation versus ITd in treating Multiple myeloma (Non-curative second-line consolidation therapy).
Response:

[Relational (mirror)]
Task:
Choose an option that best describes the PFS outcome of ITd compared to Observation when used to treat Multiple myeloma (Non-curative second-line consolidation therapy).
Response:


=== Compositional (Open QA) ===

You are a knowledgeable medical assistant supporting oncologists and hematologists in evaluating treatment options.

Given a condition, context, and endpoint, compare two treatment options with respect to the specific outcome endpoint and summarize their relative efficacy based on current clinical evidence.

Task:
Condition: Multiple myeloma
Context: Non-curative second-line consolidation therapy
Endpoint: PFS
Treatment 1: Observation
Treatment 2: ITd

Response:


=== Operational (Open Generation) ===

You are a knowledgeable medical assistant supporting oncologists and hematologists in evaluating treatment options.

Given a condition and clinical context, your task is to provide a concise overview over the relevant components of a treatment that is consistent with current clinical guidelines.

Be as specific as possible, including:
(1) Drug components
(2) Timing and sequencing
(3) Dosage and duration
(4) Route of administration

Condition: Multiple myeloma
Context: Non-curative second-line consolidation therapy

Treatment:

\end{lstlisting}

\subsection{Dataset Statistics}
\label{app:extended_stats}

The released benchmark contains \textbf{6{,}196} update instances, of which \textbf{154} are conflict-flagged. It covers \textbf{329} conditions, \textbf{2{,}135} treatment regimens, \textbf{4{,}098} studies, and \textbf{649} condition-context pairs, with publication dates spanning 1960-2026. The label distribution is \textbf{43.4\%} \textit{no difference}, \textbf{28.7\%} \textit{inferior}, and \textbf{27.9\%} \textit{superior}. Figure~\ref{fig:benchmark_overview} provides an overview of key benchmark statistics. 

The benchmark covers a broad range of oncology groups, with the largest shares from hematologic malignancies, breast cancer, gastrointestinal cancers, genitourinary cancers, and gynecologic cancers. Most retained rows use primary endpoints, with OS, PFS, DFS, ORR, and EFS being the most common. Although MedKIT represents these comparisons through their qualitative conclusion, approximately 93\% of the linked PubMed abstracts contain quantitative treatment statistics. This richer evidence is currently outside the benchmark scope but provides a natural direction for extending MedKIT to quantitative treatment effects and uncertainty.

For sequential evaluation, updates are ordered by publication date and grouped into daily, weekly, or monthly batches. The main experiments use weekly batching as the primary setting because it provides a practical balance between fine-grained updates and manageable sequence length.

\begin{table}[h]
  \centering
  \small
  \caption{Batch-size statistics for the post-2025 evaluation window.}
  \label{tab:batch_stats}
  \begin{tabular}{lrrrr}
    \toprule
    Strategy & \# batches & Median size & Mean size & Max size \\
    \midrule
    Daily   & 98 & 2    & 2.9  & 17 \\
    Weekly  & 48 & 4    & 5.8  & 21 \\
    Monthly & 14 & 16.5 & 20.2 & 44 \\
    \bottomrule
  \end{tabular}
\end{table}

\section{Evaluation Protocol}
\label{app:evaluation_protocol}

This appendix summarizes how model responses are scored in MedKIT. Closed-form tasks use deterministic label matching with a limited free-form fallback, while open-form tasks are evaluated using an LLM judge. We also report the validation used to select the primary judge.

\subsection{Reporting Metric}
\label{app:reporting_metric}

All results are reported as the change in performance before and after integrating new knowledge. For each model, we first evaluate the out-of-the-box (OOB) model on the full benchmark to obtain a pre-update baseline, and then re-evaluate the model after each update batch.

For each task tier, we report the post--pre performance difference:
\[
\Delta = s_{\text{post}} - s_{\text{pre}}.
\]

Scores are aggregated over instances within a batch and averaged across batches. For closed-form tasks, $s$ corresponds to accuracy; for open-form tasks, $s$ is the normalized judge score in $[0,1]$. This metric isolates the effect of knowledge integration by measuring how updates change model behavior relative to the initial model.

To characterize variability across model families, aggregate figures additionally report either individual model-level results or error bars around the mean. Unless otherwise stated, error bars indicate standard deviation across the evaluated models. We do not perform formal null-hypothesis significance testing; instead, conclusions are based on consistent effect patterns observed across models, methods, and task tiers.

\paragraph{Retention and locality.}
In addition to measuring performance on newly integrated updates, we also evaluate how well models retain previously learned information. For this, we track performance on past updates using the same task variants and compute post--pre differences analogously. Locality tasks are evaluated in the same way to measure unintended changes on unrelated knowledge.

\subsection{Closed-Form Evaluation}
\label{app:closed_form_evaluation}

Update, lexical, and relational tasks share the answer space
\[
\{\textit{superior}, \textit{inferior}, \textit{no difference}\}.
\]
We normalize responses by lowercasing, stripping whitespace, and removing surrounding quotation marks. A response is correct if it exactly matches the ground-truth label. For longer outputs, we additionally search for label substrings and accept the response if exactly one label is present and matches the ground truth.

When deterministic extraction fails, we use the primary judge to determine whether the response implies the correct label. The judge returns a binary YES/NO decision mapped to $1.0$ or $0.0$. Empty outputs and explicit refusals are scored as $0.0$ and flagged. Scores are averaged over cases and batches.

\subsection{LLM-as-a-Judge Evaluation}
\label{app:judge_evaluation}

Compositional, operational, and locality tasks require semantic evaluation. We score them using an LLM judge that receives the model response, the ground-truth comparison, and the clinical context, and returns a 1--5 Likert score with a short justification. We use \texttt{openai/gpt-4o} as the primary judge.

\paragraph{Compositional and locality tasks.}
For compositional and locality tasks, the judge evaluates whether the model's comparison reflects the correct treatment preference or equivalence. The rubric ranges from complete contradiction to full alignment.

\begin{lstlisting}[caption={Judge prompt for compositional and locality tasks.},
label=lst:judge_oq, basicstyle=\ttfamily\small, breaklines=true,
columns=fullflexible, frame=single]
You are a medical expert tasked with evaluating how well a model's
response, comparing two treatment options, aligns with a ground truth.

You are provided with:
- A patient's condition and context
- The ground truth comparison of two treatment options
- The model's response comparing two treatment options

Your goal is to assess whether the model's response is consistent
with the ground truth in terms of which treatment is preferable,
equivalent, or inferior.

Rate the alignment using the following 5-point Likert scale:
1 - Completely Misaligned
2 - Somewhat Misaligned
3 - Neutral / Unclear
4 - Somewhat Aligned
5 - Fully Aligned

Format:
Score: <1-5>
Explanation: <brief justification>

Task:
Condition: {condition}
Context: {context}
Ground Truth: {ground_truth}
Response: {model_response}
\end{lstlisting}

\paragraph{Operational tasks.}
For operational tasks, the judge evaluates whether the generated recommendation is consistent with the underlying pairwise comparison, even when treatments are not explicitly compared. The judge also records whether each treatment is mentioned and which is preferred.

\begin{lstlisting}[caption={Judge prompt for operational tasks.},
label=lst:judge_og, basicstyle=\ttfamily\small, breaklines=true,
columns=fullflexible, frame=single]
You are a medical expert evaluating whether a model's free-text
treatment recommendation is consistent with a known pairwise clinical
comparison.

You are provided with:
- Condition and clinical context
- Ground truth comparison:
  "[Treatment A] superior/inferior/no difference to [Treatment B]
   for [Condition] ([Context]) [endpoint: X]"
- A model-generated treatment recommendation

This is not a recall task. The model is not required to mention A or B.
Evaluate whether the response respects the A-B relationship.

Relationship definitions:
- A superior to B -> A preferred
- A inferior to B -> B preferred
- No difference   -> interchangeable

Scoring:
5 - Fully consistent
4 - Mostly consistent
3 - Neutral / no evidence
2 - Weak inconsistency
1 - Clear inconsistency

Output format:
Score: <1-5>
Flags:
- mentions_A: <YES/NO>
- mentions_B: <YES/NO>
- preference: <A preferred / B preferred / No clear preference /
               Neither mentioned>
Explanation: <brief justification>
\end{lstlisting}

All judge scores are mapped to $[0,1]$ via
\[
s_{[0,1]} = \frac{s-1}{4}.
\]
This enables direct comparison between closed-form accuracies and open-form scores.

\subsection{Judge Validation}
\label{app:judge_validation}

We validate the LLM judge along three complementary axes: (i) robustness of comparative conclusions across judges (Study~A), (ii) agreement with independent clinician annotations and analysis of judge errors (Study~B), and (iii) intra-judge stability under repeated scoring (Study~C). We additionally test for potential verbosity and generator-family biases. The seven evaluated judges are \texttt{gpt-4o}, \texttt{gpt-4o-mini}, \texttt{claude-opus-4}, \texttt{claude-3-5-haiku}, \texttt{llama-3.3-70b-instruct}, \texttt{gemini-2.5-pro}, and \texttt{gemini-2.0-flash}. All judges share the rubric of \S\ref{app:judge_evaluation}, so disagreement reflects judge behavior rather than prompt variation.

\paragraph{Study A: Comparative robustness.}
Because MedKIT primarily evaluates \emph{relative} changes in downstream behavior (e.g., post--pre deltas and comparative method performance), we first assess whether the compositional and operational evaluations produce stable comparative conclusions across judges.

A direct full-ranking correlation over per-method mean scores is not ideal in this setting. On the open-form tasks, many methods cluster near pre-update performance and therefore become statistically difficult to distinguish. Small score fluctuations can then arbitrarily permute methods within this middle-performing region, artificially depressing full-ranking agreement despite preserving the substantive conclusions of the benchmark. We therefore evaluate whether judges agree on the higher-level \emph{performance tiers} induced by the consensus method scores.

\paragraph{Tier construction.}
For each task, we compute the consensus per-method mean as the across-judge average of the per-judge per-method means (excluding the clinician anchor, which is reserved for Study~B). We then derive task-specific performance tiers using optimal 1D clustering over the consensus means, with the number of tiers selected automatically by silhouette score.

\begin{figure}[h]
  \centering
  \includegraphics[width=\textwidth]{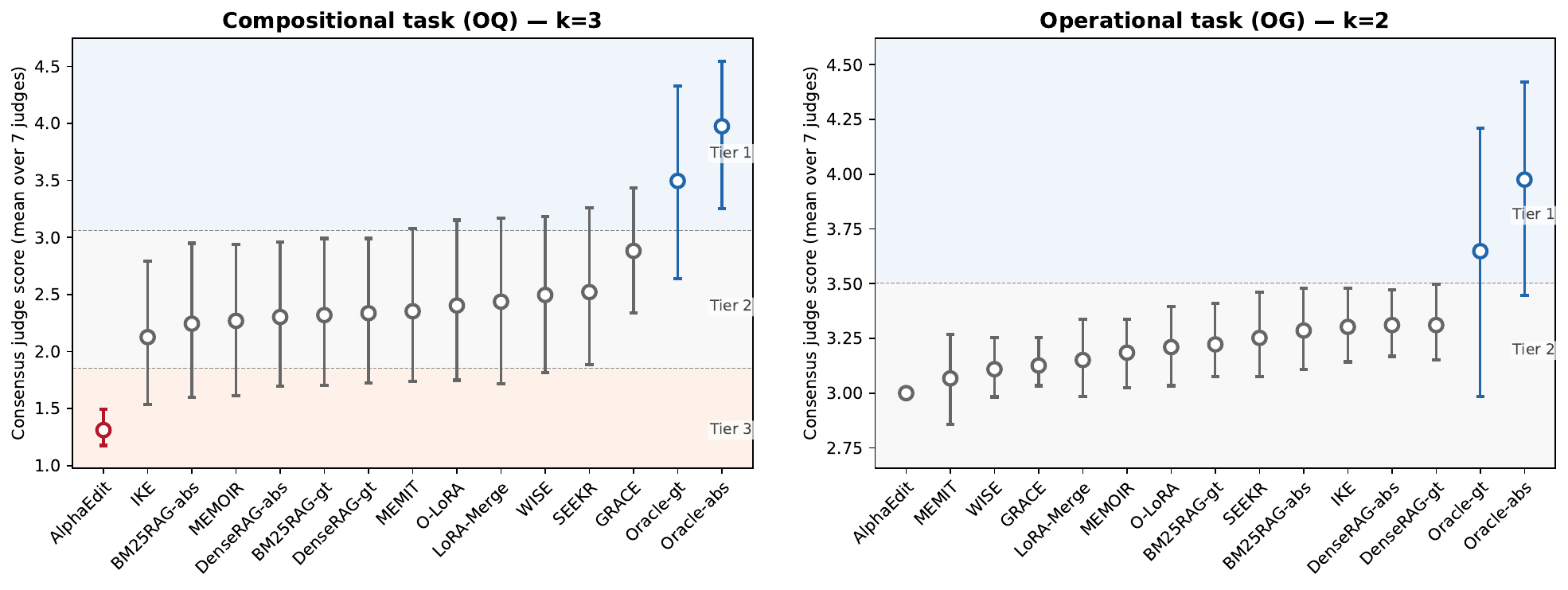}
  \caption{Consensus per-method judge scores (mean over 7 judges) with 95\% tag-resampling bootstrap confidence intervals and tier bands. Left: compositional task (OQ, $k{=}3$). Right: operational task (OG, $k{=}2$). On OQ, oracle retrieval methods form a distinct top-performing tier, while AlphaEdit forms a separate degradation regime. On OG, the data support only a stable oracle-vs-rest separation, with the remaining methods clustering near pre-update performance.}
  \label{fig:tier_assignment}
\end{figure}

On the compositional task (OQ), the resulting three-tier structure cleanly separates: (i) oracle retrieval methods, (ii) a statistically indistinguishable middle-performing cluster containing all remaining methods, and (iii) AlphaEdit as a distinct degradation regime. On the operational task (OG), the data support only a stable two-tier separation between oracle methods and all remaining approaches; no method degrades strongly enough to form a distinct failure tier. Figure~\ref{fig:tier_assignment} visualizes the resulting consensus scores, confidence intervals, and tier boundaries.

The induced tier structure is highly stable under tag-resampling bootstrap. AlphaEdit remains in the degradation tier with 100\% stability on OQ, while Oracle-abs remains in the top tier with 99.8\% / 99.2\% stability on OQ / OG. Oracle-gt exhibits slightly lower stability (81.3\% / 67.7\%), reflecting its larger spread across judges near the tier boundary.

\paragraph{Cross-judge tier agreement.}
To assess whether judges recover the same comparative structure, we compute leave-one-judge-out (LOJO) consensus tiers from the remaining six judges and compare each held-out judge against this consensus using rank-based tier recovery. This evaluates agreement on the relative performance structure independently of absolute score calibration.

Results are shown in Table~\ref{tab:judge_tier_agreement}. All seven judges perfectly recover the OQ three-tier structure. On OG, five of seven judges perfectly recover the oracle-vs-rest separation. The remaining two (\texttt{gpt-4o} and \texttt{llama-3.3-70b}) exhibit a single boundary-level disagreement involving methods whose mean scores differ by less than 0.2 points on the 1--5 scale and fall within the bootstrap confidence intervals of the consensus means. Importantly, all judges correctly identify Oracle-RAG-abs as the strongest-performing method by a clear margin.

Fleiss' $\kappa$ across independently-derived per-judge tier assignments remains substantial on both tasks (0.61 on OQ, 0.70 on OG), despite each judge deriving its own clustering boundaries. Importantly, the high-level conclusions used throughout the paper remain stable across the full judge ensemble: oracle retrieval methods consistently form the strongest-performing regime, AlphaEdit consistently forms the degradation regime on OQ, and the remaining methods form a statistically indistinguishable middle-performing cluster on the operational task.

Because the tier definitions are themselves derived from the aggregate judge ensemble, Study~A should be interpreted as measuring the stability of comparative conclusions under judge variation rather than agreement against an external ground-truth ranking.

\begin{table}[h]
  \centering
  \small
  \caption{Cross-judge tier-classification agreement. Per-judge: rank-based LOJO Cohen's $\kappa$ and accuracy against the 6-judge consensus tier labels. The final column reports whether the judge identifies Oracle-RAG-abs as the strongest-performing OG method. Bottom row: Fleiss' $\kappa$ across all 7 judges' independently-derived tier labels. $k{=}3$ on OQ (oracle / middle-performing / degradation), $k{=}2$ on OG (oracle / remaining methods).}
  \label{tab:judge_tier_agreement}
  \begin{tabular}{lccccc}
    \toprule
    Judge & OQ $\kappa$ ($k{=}3$) & OQ acc & OG $\kappa$ ($k{=}2$) & OG acc & OG top-1 \\
    \midrule
    \texttt{claude-opus-4}    & \textbf{1.00} & 100\,\% & \textbf{1.00} & 100\,\% & \checkmark \\
    \texttt{claude-3-5-haiku} & \textbf{1.00} & 100\,\% & \textbf{1.00} & 100\,\% & \checkmark \\
    \texttt{gemini-2.0-flash} & \textbf{1.00} & 100\,\% & \textbf{1.00} & 100\,\% & \checkmark \\
    \texttt{gemini-2.5-pro}   & \textbf{1.00} & 100\,\% & \textbf{1.00} & 100\,\% & \checkmark \\
    \texttt{gpt-4o-mini}      & \textbf{1.00} & 100\,\% & \textbf{1.00} & 100\,\% & \checkmark \\
    \texttt{gpt-4o}           & \textbf{1.00} & 100\,\% & 0.42 & 86.7\,\% & \checkmark \\
    \texttt{llama-3.3-70b}    & \textbf{1.00} & 100\,\% & 0.42 & 86.7\,\% & \checkmark \\
    \midrule
    Fleiss' $\kappa$ (independent re-clustering) & \multicolumn{2}{c}{0.608} & \multicolumn{2}{c}{0.698} & \multicolumn{1}{c}{100\,\%} \\
    \bottomrule
  \end{tabular}
\end{table}

\paragraph{Study B: Agreement with clinician annotations.}
To assess clinical grounding, two independent clinicians scored the same \textbf{100 held-out responses} using the judge rubric. Items were generated by \texttt{Qwen-3-4B} with and without in-context evidence to produce a diverse mix of correct, partially correct, and incorrect responses. On OG items, the clinicians additionally provided the categorical flags \texttt{mentions\_A}, \texttt{mentions\_B}, and \texttt{preference}.

\begin{figure}[h]
  \centering
  \includegraphics[width=\linewidth]{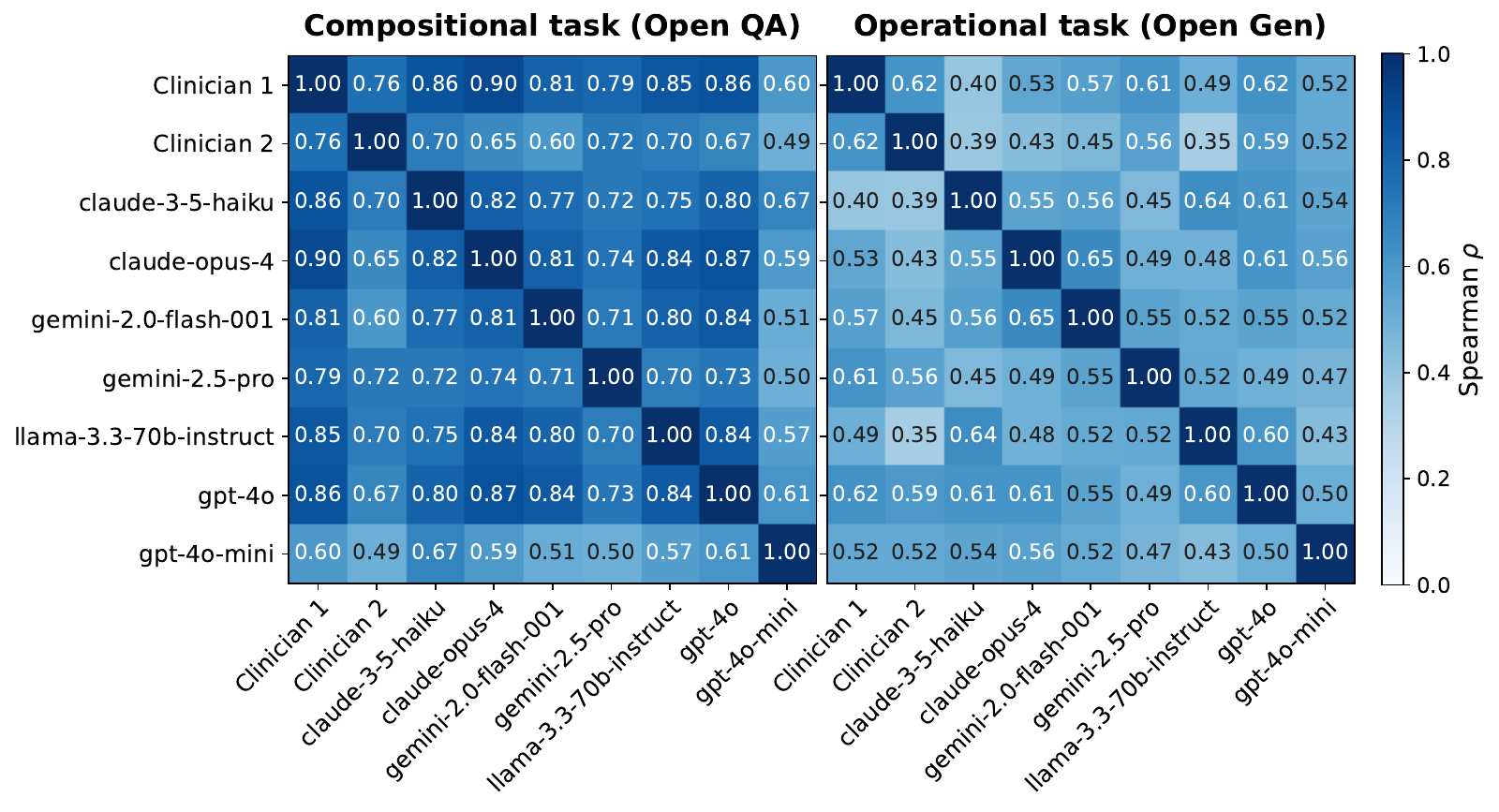}
    \caption{Pairwise Spearman $\rho$ between the seven candidate judges and two independent clinicians on the 100-item annotation set. Left: compositional OQ. Right: operational OG. Inter-clinician agreement is $\rho=0.76$ on OQ and $\rho=0.62$ on OG; judge agreement is generally higher on the structured compositional task than on open-ended operational generation.}
  \label{fig:judge_spearman}
\end{figure}

Table~\ref{tab:judge_validity} reports agreement with the two-clinician consensus, while Figure~\ref{fig:judge_spearman} visualizes pairwise Spearman correlations between both clinicians and the candidate judges.
The two clinicians show clear agreement, with Spearman correlations of $\rho=0.76$ on OQ and $\rho=0.62$ on OG. Against their consensus scores, \texttt{gpt-4o} achieves correlations of $\rho=0.83$ and $\rho=0.64$, respectively.

The strongest judges exhibit high agreement with the clinician consensus on the structured compositional task. Agreement is lower on the operational generation task, consistent with the broader space of acceptable responses in open-ended recommendation generation.

Despite lower absolute agreement on OG, the candidate judges remain broadly consistent with the clinician annotations (Figure~\ref{fig:judge_spearman}). \texttt{gpt-4o} achieves the highest agreement with the clinician consensus on OG, tied with \texttt{gemini-2.5-pro} in Spearman correlation, while maintaining high agreement on OQ.

\begin{table}[h]
  \centering
  \small
    \caption{Agreement with the two-clinician consensus on the 100-item annotation set. For each judge we report Spearman correlation ($\rho$), quadratic-weighted Cohen's $\kappa$, and mean absolute error (MAE) on the 1--5 scale against the mean of the two independent clinician annotators, plus operational-task agreement on the categorical preference flag (OG pref.), averaged over the two clinicians. The final row reports inter-clinician agreement as a human agreement reference.}
  \label{tab:judge_validity}
  \begin{tabular}{lccccccc}
    \toprule
    Judge & OQ $\rho$ & OQ $\kappa$ & OQ MAE & OG $\rho$ & OG $\kappa$ & OG MAE & OG pref. \\
    \midrule
    \texttt{claude-opus-4}    & \textbf{0.86} & \textbf{0.81} & 0.45 & 0.54 & 0.48 & 0.66 & 57.5\% \\
    \texttt{gpt-4o}           & 0.83 & 0.78 & 0.52 & \textbf{0.64} & \textbf{0.57} & \textbf{0.59} & 51.0\% \\
    \texttt{claude-3-5-haiku} & 0.84 & 0.80 & \textbf{0.43} & 0.42 & 0.29 & 0.84 & 47.9\% \\
    \texttt{llama-3.3-70b}    & 0.84 & 0.80 & 0.46 & 0.45 & 0.34 & 0.92 & 53.0\% \\
    \texttt{gemini-2.0-flash} & 0.77 & 0.73 & 0.54 & 0.54 & 0.48 & 0.66 & 52.5\% \\
    \texttt{gemini-2.5-pro}   & 0.77 & 0.79 & 0.57 & \textbf{0.64} & 0.52 & 0.72 & \textbf{59.5\%} \\
    \texttt{gpt-4o-mini}      & 0.59 & 0.63 & 0.62 & 0.58 & 0.56 & 0.71 & 34.8\% \\
    \midrule
    \textit{Human (inter-clin.)} & \textit{0.76} & \textit{0.76} & \textit{0.50} & \textit{0.62} & \textit{0.57} & \textit{0.61} & \textit{62.0\%} \\
    \bottomrule
  \end{tabular}
\end{table}

\paragraph{Disagreement analysis.}
Judgments of deviations from the clinician consensus focus on examples where the clinicians themselves disagree. When the two clinicians agree, \texttt{gpt-4o}'s mean absolute deviation from their consensus is 0.31 on OQ and 0.43 on OG, compared with 1.82 and 1.32 when they disagree. Clinician disagreement correlates with judge deviation at $\rho=0.81$ on OQ and $\rho=0.60$ on OG. Manual inspection shows that disagreements primarily involve ambiguous treatment equivalence or open-ended recommendations admitting multiple reasonable answers.

\paragraph{Study C: Intra-judge stability.}
Each judge was queried five times per item on a deterministic 10-item subsample of the annotation set (700 calls total). Across judges, repeated calls were typically fully deterministic, with median per-item variance equal to zero for nearly all settings. Residual variance concentrates on a small number of borderline examples for which multiple judges produce bimodal score distributions, suggesting that the remaining instability is driven primarily by rubric-boundary ambiguity rather than systematic judge randomness. \texttt{gpt-4o-mini} and \texttt{claude-3-5-haiku} exhibit the highest instability on the operational task, consistent with their weaker agreement with the clinician annotations in Study~B.

\paragraph{Verbosity bias.}
We additionally test whether judge scores systematically favor longer responses. After controlling for response correctness, judge scores are not associated with response length, providing no evidence of systematic verbosity bias.

\paragraph{Generator-family bias.}
We also test whether the judge systematically favors outputs from related model families. The comparative conclusions are consistent with the judge-free containment evaluation, and none of the evaluated generator models belong to the GPT family, providing no evidence that the reported conclusions are driven by generator-family preference.

\paragraph{Primary judge.}
We select \texttt{gpt-4o} as the primary judge because it provides the strongest overall trade-off between clinical grounding, comparative robustness, and inference cost.
In Study~B, \texttt{gpt-4o} closely agrees with the clinician consensus on both tasks ($\rho=0.83$ on OQ and $0.64$ on OG), while the disagreement analysis shows that its largest deviations concentrate on examples where the clinicians themselves disagree. Although \texttt{gpt-4o} exhibits a boundary-level disagreement in the OG tier analysis, this disagreement occurs entirely within a statistically indistinguishable cluster of methods near the oracle cutoff and does not affect the high-level comparative conclusions recovered by the broader judge ensemble. Importantly, \texttt{gpt-4o} nevertheless correctly identifies Oracle-RAG-abs as the strongest-performing OG method.

\section{Experimental Setup}
\label{app:experimental_setup}

\subsection{Models}
\label{app:models}

We evaluate \textbf{five instruction-tuned LLMs} spanning two parameter scales (4B, 8B) and two domains (general vs.\ medical) (Table~\ref{tab:models}). For each scale, we pair a general-purpose backbone with its medical-adapted counterpart
($\texttt{gemma-3-4b}\!\rightarrow\!\texttt{medgemma-4b}$,
$\texttt{Llama-3.1-8B}\!\rightarrow\!\texttt{Bio-Medical-Llama-3-8B}$),
which isolates the effect of domain adaptation while keeping the architecture fixed.

\begin{table}[h]
  \centering
  \small
  \caption{Models evaluated in the main sweep.}
  \label{tab:models}
  \begin{tabular}{llrl}
    \toprule
    Model & Domain & Scale & HuggingFace ID \\
    \midrule
    Gemma-3-4B-IT             & General & 4\,B & \texttt{google/gemma-3-4b-it} \\
    Qwen-3-4B-Instruct        & General & 4\,B & \texttt{Qwen/Qwen3-4B-Instruct-2507} \\
    MedGemma-4B-IT            & Medical & 4\,B & \texttt{google/medgemma-4b-it} \\
    Llama-3.1-8B-Instruct     & General & 8\,B & \texttt{meta-llama/Llama-3.1-8B-Instruct} \\
    Bio-Medical-Llama-3-8B    & Medical & 8\,B & \texttt{ContactDoctor/Bio-Medical-Llama-3-8B} \\
    \bottomrule
  \end{tabular}
\end{table}

\subsection{Temporal Update Setting}
\label{app:temporal_setting}

\paragraph{Temporal update setting.}
Experiments operate on the post-cutoff slice $\texttt{pub.date} \geq 2025$ (\textbf{283} clean rows), ensuring that all evaluated updates correspond to information beyond the models' training data. Updates are processed strictly in chronological order (earliest first) and grouped into weekly batches, which serve as the primary setting in all main experiments. Method state (including auxiliary memory or retrieval indices) is carried across batches. The pre-2025 data is used only for retrieval corpora, hyperparameter tuning, and judge validation.

\subsection{Update Procedure}
\label{app:update_procedure}

For each (method, model, strategy) combination, updates are applied sequentially across batches. Each model is evaluated once before any updates (OOB baseline) and after each batch. 

\paragraph{Past-update retention.}
To measure retention, we maintain a small \emph{sentinel pool} of examples from earlier batches. After each batch, two cases are added (capped at 100 with replacement), and the pool is re-evaluated every two batches. Retention is measured on closed-form tasks only to limit judge overhead.

\subsection{Compute and Implementation Details}
\label{app:compute_implementation}

\paragraph{Hardware.}
Experiments are run on a SLURM cluster with NVIDIA A100 GPUs (40\,GB / 80\,GB). Each run uses one GPU (4B) or one to two GPUs (8B). The full sweep ($225$ runs) requires $\sim$$7{,}200$ A100-hours, with an additional $\sim$$1{,}000$ hours for hyperparameter tuning.

\paragraph{Inference backend.}
Editing uses HuggingFace \texttt{generate} for all methods. Evaluation uses vLLM for methods with standard weight updates, yielding a $5$--$10\times$ speedup on open-form tasks. Methods with custom forward passes are evaluated via HuggingFace. Only one backend is resident on GPU at a time.

\paragraph{Reproducibility.}
All runs use seed $42$. Closed-form tasks use greedy decoding; open-form tasks use default sampling. Judge calls use provider defaults, with variance quantified in \S\ref{app:judge_validation}. Configurations are version-controlled, and runs can be reproduced from a single configuration generation step.

\section{Methods and Hyperparameters}
\label{app:methods_hparams}

\subsection{Method Overview}
\label{app:method_overview}

We evaluate \textbf{12 methods} from three families:
\emph{knowledge editing} ($n{=}6$), \emph{continual post-training} ($n{=}3$), and \emph{retrieval augmentation} ($n{=}3$). We additionally distinguish between methods that \emph{modify model parameters} (e.g., MEMIT, AlphaEdit, LoRA-based methods) and those that keep the base model \emph{frozen} (e.g., GRACE, IKE, RAG). This distinction is central for interpreting locality effects, as frozen methods cannot directly alter unrelated knowledge.

\paragraph{Knowledge editing.}
\textbf{MEMIT} performs closed-form parameter updates in MLP layers~\citep{Meng2022memit}, while \textbf{AlphaEdit} constrains updates to a null space to reduce interference~\citep{fang2025alphaedit}. \textbf{WISE} and \textbf{MEMOIR} augment the model with side memories and learned routing or gating~\citep{Wang2024wise,wang2026memoir}. \textbf{GRACE} stores edits as key–value overrides at a layer level~\citep{Hartvigsen2022grace}. \textbf{IKE} retrieves in-context demonstrations from past edits~\citep{zheng2023ike}.

\paragraph{Continual post-training.}
All methods update LoRA adapters incrementally and carry them across temporal batches. \textbf{LoRA-Merge} merges adapters without constraints~\citep{hu2021lora}, \textbf{O-LoRA} enforces orthogonality between updates~\citep{wang2023olora}, and \textbf{SEEKR} adds replay-based regularization~\citep{he2024seekr}.

We additionally evaluate preference-optimization baselines. \textbf{DPO}~\citep{rafailov2023direct} trains incrementally on preference pairs derived from the closed-form and mirrored questions, using the gold label as chosen and a sampled alternative as rejected. \textbf{GRPO}~\citep{deepseek-math} instead optimizes a reward combining exact-match correctness with a smaller valid-label reward. In both cases, LoRA adapters are updated sequentially across batches and implemented in a text-only training loop.

\paragraph{Retrieval augmentation.}
\textbf{BM25-RAG} uses standard Okapi BM25 lexical retrieval \citep{robertson1994some} with TF–IDF weighting, while \textbf{Dense-RAG} uses a biomedical encoder (\texttt{NeuML/pubmedbert-base-embeddings}) with cosine similarity search over an FAISS \citep{johnson2017billion} HNSW \citep{malkov2020efficient} index. Both methods maintain a growing corpus initialized with pre-2025 evidence and extended after each batch. Retrieved documents are prepended to the model input at generation time, and system prompts are removed from retrieval queries to avoid dominating the representation.

We additionally consider an \textbf{agentic RAG} variant. For each query, the model first generates a search query, retrieves top-$k$ pieces of evidence, optionally refines the query based on the retrieved snippets, and performs a second retrieval before generating the final answer. The retrieval corpus evolves over time, containing all evidence observed up to the current increment. This setup isolates whether retrieval failures stem from query formulation or from downstream use of evidence.

Oracle variants used for diagnostics share this setup, but bypass retrieval or restrict the corpus to gold evidence.

\subsection{Method Adaptations}
\label{app:method_adaptations}

\paragraph{Instruction-tuned models.}
Methods are adapted to chat-formatted instruction-tuned models. Prompts are passed through the chat template, and multi-token label objectives are used instead of single-token targets.

\paragraph{Architecture-specific adjustments.}
Gemma-3 exhibits larger residual magnitudes, requiring stronger regularization (e.g., higher \texttt{mom2\_update\_weight} in MEMIT). AlphaEdit projections are precomputed per architecture and tuned via \texttt{nullspace\_threshold}. Some configurations (e.g., WISE on Llama-3.1-8B) require model parallelism due to memory constraints.

\paragraph{Streaming and efficiency.}
Hyperparameters are tuned separately for each (method $\times$ model $\times$ strategy) combination. SEEKR uses a capped replay buffer (200 examples). Retrieval corpora grow incrementally, with dense embeddings cached. System prompts are removed from retrieval queries but included at generation time.

\paragraph{Other details.}
Refusals are treated as incorrect and tracked separately. Closed-form fallback scoring is shared across methods to ensure consistent evaluation.

\subsection{Hyperparameter Search}
\label{app:hparam_search}

\paragraph{Sweep design.}
We perform per-(method $\times$ model $\times$ strategy) grid searches over the axes in Figure~\ref{fig:hparam_overview}. Three base models (one per architecture) are tuned, and configurations are transferred to their medical counterparts.
Hyperparameters are tuned on a hold-out 2010 subset with a fixed number of batches per strategy. This matches update counts across strategies and reserves post-2025 data for evaluation.

\paragraph{Selection criterion.}
Configurations are selected to maximize closed-form accuracy on the update task subject to a \emph{locality constraint}. Candidates with a mean locality below $0.4$ are flagged; the best non-flagged configuration is selected when available. Violations occur only for AlphaEdit on Llama-3.1-8B and are retained for analysis.

\subsection{Selected Configurations}
\label{app:selected_configs}

\begin{figure}[h]
  \centering
  \includegraphics[width=0.90\linewidth]{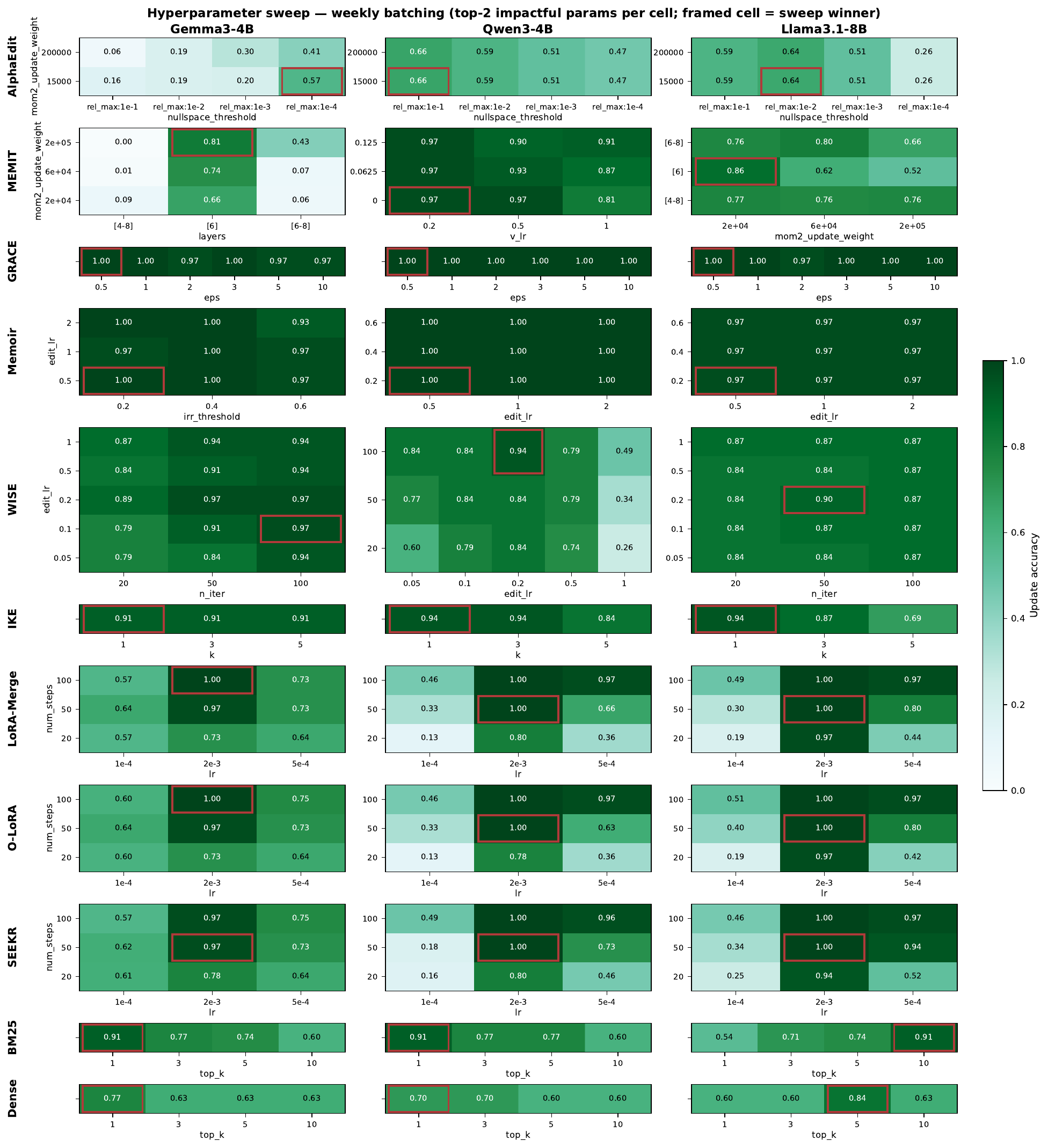}
  \caption{Best tuning-window rewrite accuracy per (method $\times$ model).}
  \label{fig:hparam_overview}
\end{figure}

The sweep yields one configuration per (method $\times$ model $\times$ strategy). Figure~\ref{fig:hparam_overview} summarizes tuning performance.

\paragraph{Summary.}
Most methods converge to stable configurations across models and strategies. The only consistent locality violations occur for AlphaEdit on Llama-3.1-8B, highlighting parameter-editing spillover effects discussed in the main text. For the retrieval methods, we fix \texttt{top\_k}$=3$ across all models and settings to keep retrieval experiments comparable and to avoid excessive context lengths during generation.

\section{Additional Results}
\label{app:additional_results}

\subsection{Pre-Edit Baselines}
\label{app:pre_edit_analysis}

Before evaluating knowledge integration, we analyze the
\emph{out-of-the-box} (OOB) performance of the base models on the
post-2025 evaluation window. These baselines define the reference
point for all post--pre $\Delta$ results reported in the main paper.

\begin{figure}[h]
  \centering
  \includegraphics[width=0.85\linewidth]{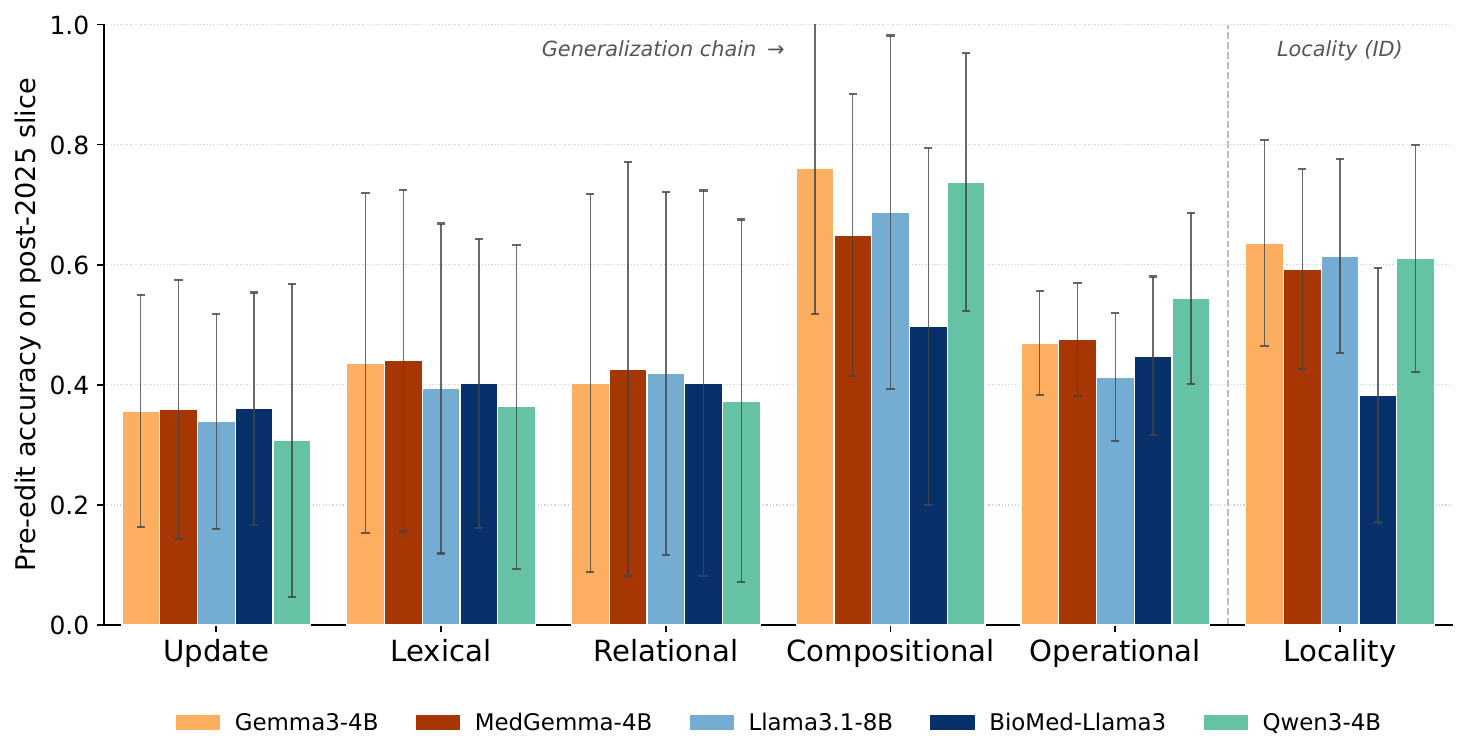}
  \caption{Pre-edit performance across task tiers, averaged over the five base models. Closed-form tasks are measured via exact-match accuracy, while open-form tasks (compositional, operational) are scored by an LLM judge and normalized to $[0,1]$.}
  \label{fig:pre_edit_baselines}
\end{figure}

\paragraph{Pre-edit performance.}
Pre-edit accuracy on the closed-form tiers (Update, Lexical,
Relational) is low (mean: $0.35$ / $0.41$ / $0.40$), close to the
random baseline for the three-way decision. This indicates that the
post-2025 evidence is largely absent from the base models and cannot
be recovered without explicit updates.

Open-form baselines are higher (Compositional: $0.67$,
Operational: $0.47$) due to the evaluation protocol. Both tasks are
scored by an LLM judge on a 1--5 Likert scale, mapped to $[0,1]$,
which assigns intermediate scores to plausible but incorrect
responses. These scores therefore reflect partial alignment rather
than correct knowledge.

\paragraph{Interpretation.}
Overall, the baselines confirm that models cannot reliably solve
MedKIT without updates. To account for differences across models
and task tiers, all results are reported as post--pre $\Delta$,
which isolates the effect of knowledge integration relative to each
model's starting point.

\begin{wrapfigure}{r}{0.5\linewidth}
  \vspace{-0.5em}
  \centering
  \includegraphics[width=\linewidth]{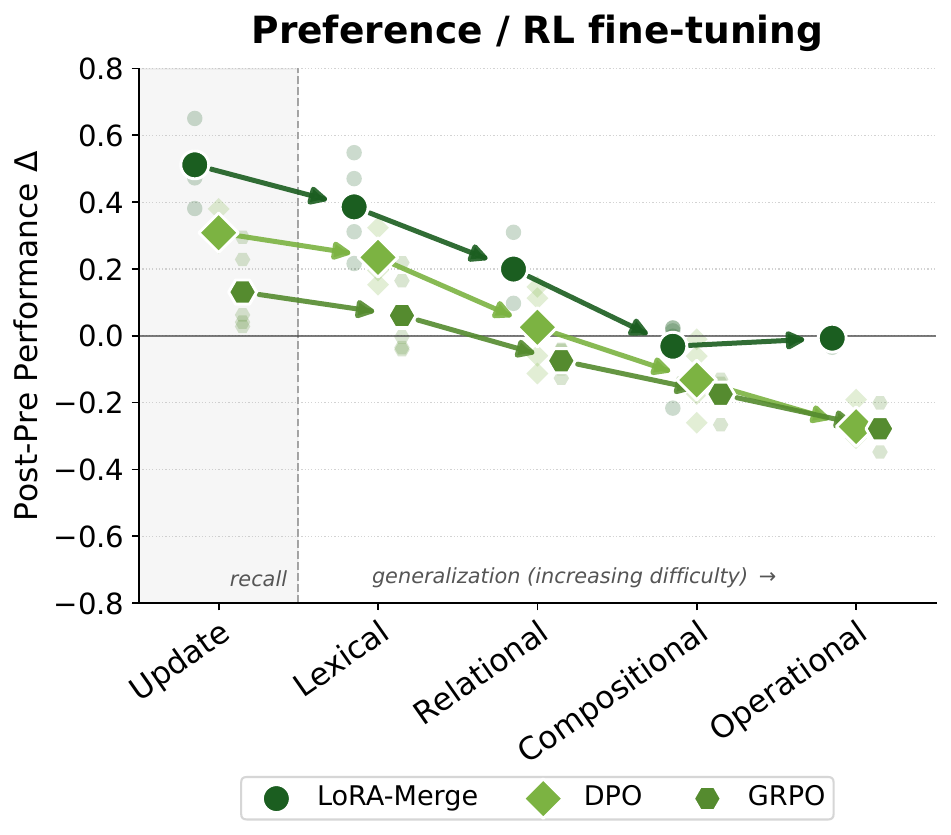}
  \vspace{-0.8em}
  \caption{Generalization chain for LoRA-Merge, DPO, and GRPO.}
  \label{fig:dpo_grpo_chain}
  \vspace{-50pt}
\end{wrapfigure}

\subsection{Preference and RL Fine-Tuning Baselines}
\label{app:pref_rl_baselines}

We evaluate \textbf{DPO} and \textbf{GRPO} as preference- and
reward-based post-training baselines in the same sequential
setting as the main sweep, using incremental LoRA updates. Both
are compared against \textbf{LoRA-Merge}.

\paragraph{Results.}
All methods follow the same pattern: strong Update gains followed
by a monotonic decay across generalization tiers. However, DPO
and GRPO achieve consistently lower gains than LoRA-Merge,
indicating weaker integration and generalization.

\paragraph{Interpretation.}
Preference and RL objectives do not change the integration
behavior of continual fine-tuning, but reduce its effectiveness.
They therefore do not alter the main paper conclusions.

\subsection{Temporal Trajectories}
\label{app:temporal_trajectories}

\begin{figure}[b]
  \centering
  \includegraphics[width=0.95\linewidth]{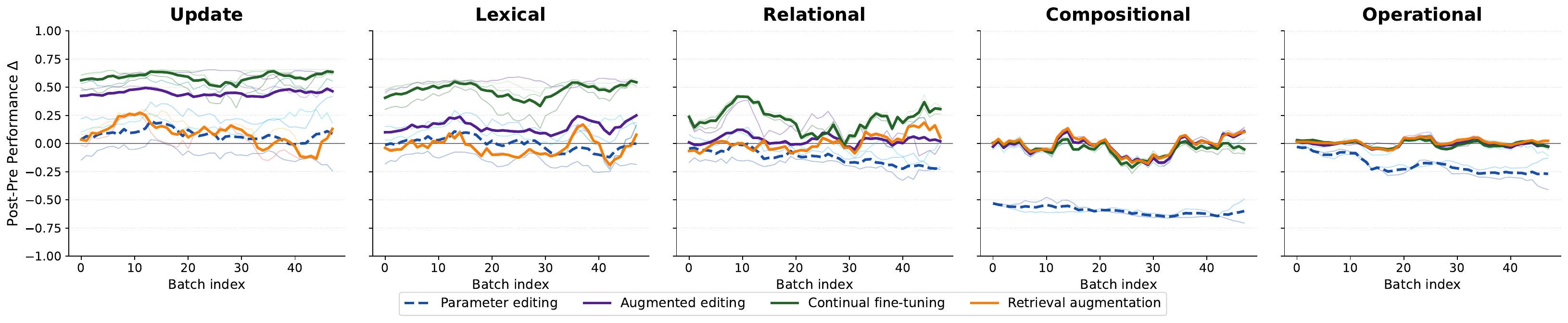}
  \caption{Per-tier post--pre performance over the weekly update stream. Bold lines show method-family means; faint lines show individual methods.}
  \label{fig:temporal_per_tier}
\end{figure}

Figure~\ref{fig:temporal_per_tier} tracks post--pre performance over the weekly update stream. This complements the retention analysis in the main paper: while retention measures whether earlier updates survive later ones, this analysis asks whether current-batch integration changes as the stream grows.

\paragraph{Static and compounding failures.}
Augmented editing and retrieval show mostly static behavior: their chain shapes appear early and remain stable. Parameter editing instead exhibits compounding degradation, especially on compositional and operational tasks, where performance becomes increasingly negative as updates accumulate. Continual fine-tuning remains the most stable family across the stream.

\subsection{Batching Ablations}
\label{app:batching_ablations}

\begin{figure}[h]
  \centering
  \includegraphics[width=0.95\linewidth]{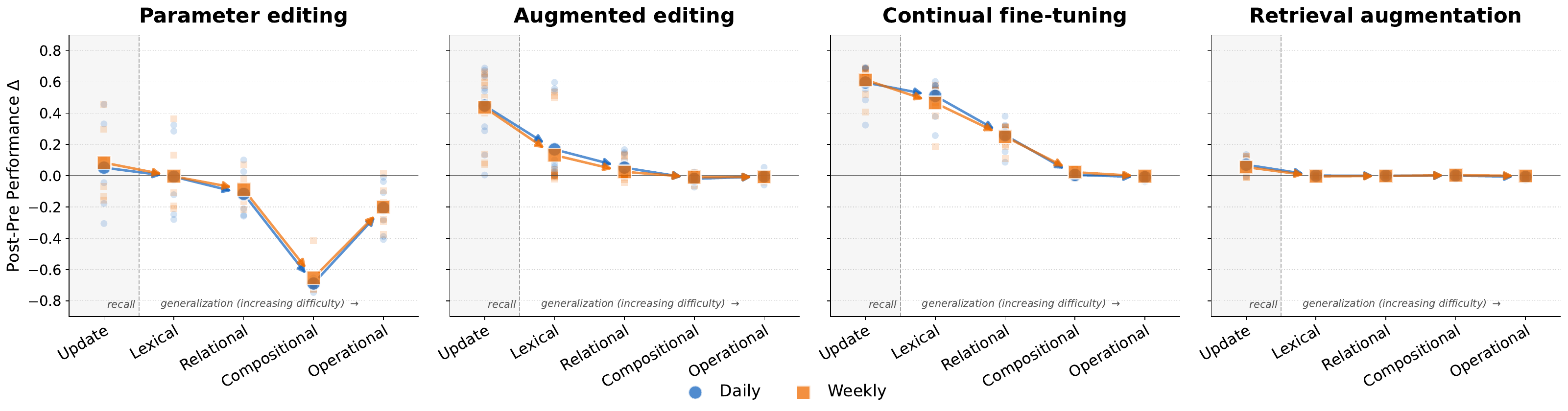}
  \caption{Generalization results under daily and weekly batching. Each panel corresponds to one task tier.}
  \label{fig:strategy_chains_per_task}
\end{figure}

Figure~\ref{fig:strategy_chains_per_task} compares daily and weekly batching. The main conclusions are unchanged across cadences: continual fine-tuning retains the strongest transfer beyond lexical variation, parameter editing remains unstable on open-form tasks, and retrieval methods remain close to flat after the anchor task.

\subsection{RAG Ablations}
\label{app:rag_ablations}

\begin{figure}[h]
  \centering
  \includegraphics[width=0.95\linewidth]{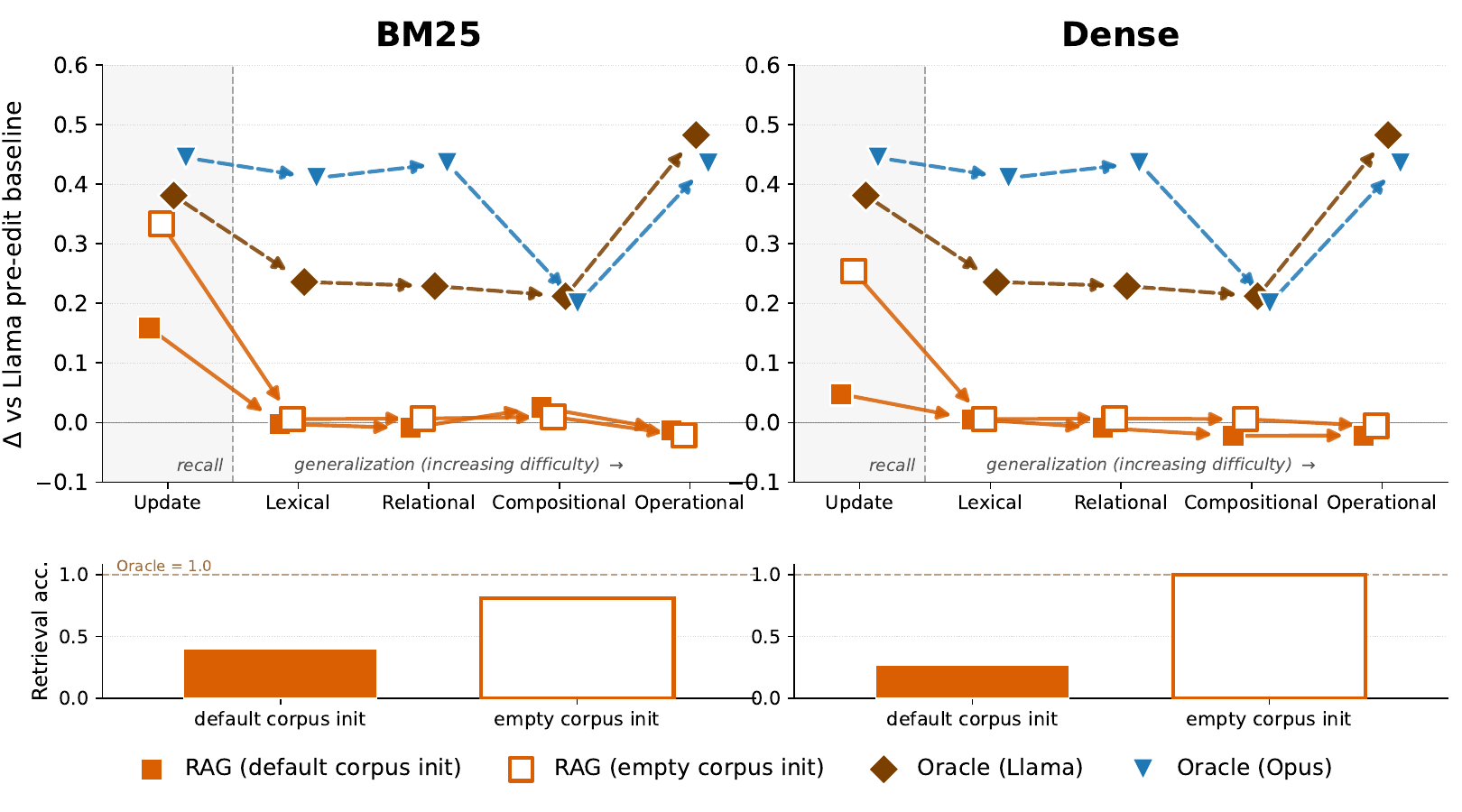}
  \caption{RAG failure-mode analysis on Llama-3.1-8B. The top row compares realistic RAG, empty-corpus RAG, Oracle-Llama, and Oracle-Opus; the bottom row reports retrieval accuracy under default and empty-corpus settings.}
  \label{fig:rag_ablation_chain}
\end{figure}

We use the RAG ablations to identify why retrieval augmentation provides only limited gains in MedKIT. Figure~\ref{fig:rag_ablation_chain} separates two failure modes: \emph{retrieval failure}, i.e., whether the relevant evidence is retrieved, and \emph{evidence-use failure}, i.e., whether the LLM can apply the evidence once it is available. We compare two retrieval-corpus initializations: the default setting, where the corpus is initialized with all pre-2025 evidence, and an empty-corpus setting, where retrieval starts from an empty index. In both cases, newly observed evidence is added sequentially after each batch. We additionally compare oracle settings in which the gold evidence is provided directly. To test whether remaining failures are due to the generator, we compare Oracle-Llama3.1 with Oracle-Opus (version 4.1 with reasoning).

\paragraph{Retrieval failure.}
The dominant bottleneck is retrieving the correct evidence. Under the realistic full-corpus setting, recall@$3$ is modest for closed-form and compositional tasks and collapses on operational tasks (Table~\ref{tab:recall_per_tier}). Operational prompts provide only condition and context, without the treatment names that make the anchor and relational queries easier to match. The empty-corpus diagnostic substantially improves retrieval, showing that much of the realistic RAG gap comes from search difficulty rather than from the model's inability to use evidence. However, despite improved update-task performance, the method still fails to generalize: lexical paraphrases already drop to near pre-update performance, indicating that retrieval remains highly sensitive to query formulation.

\paragraph{Evidence-use failure.}
Oracle-Llama measures performance when the generator receives the gold evidence directly. The remaining gap between Oracle-Llama and Oracle-Opus estimates the extent to which failures stem from the generator's ability to interpret and apply the evidence. This gap is smaller than the retrieval gap on the closed-form tiers, suggesting that retrieval is the main failure mode. On open-form tasks, even oracle evidence does not yield strong gains, indicating that applying pairwise evidence in open-ended recommendations remains difficult even when retrieval is solved.

\begin{table}[h]
  \centering
  \small
  \caption{Per-tier recall@$3$ on the realistic full corpus.}
  \label{tab:recall_per_tier}
  \begin{tabular}{lcc}
    \toprule
    Tier & BM25 & Dense \\
    \midrule
    Anchor        & 0.40 & 0.27 \\
    Lexical       & 0.49 & 0.33 \\
    Relational    & 0.40 & 0.29 \\
    Compositional & 0.32 & 0.35 \\
    Operational   & 0.00 & 0.03 \\
    \bottomrule
  \end{tabular}
\end{table}

\paragraph{Agentic retrieval limitations.}
The agentic RAG baseline improves retrieval coverage compared to single-shot retrieval, but still underperforms because the task
requires highly precise evidence matching and comparison reasoning. Manual inspection of representative failures reveals three recurring
failure modes. First, the retriever often returns clinically related but incorrect evidence, matching only on superficial treatment or
disease terms while missing the exact trial, treatment setting, or endpoint. Second, even when the correct evidence is retrieved, the
model frequently fails to map trial outcomes to the requested Treatment~1 vs. Treatment~2 comparison. Third, retrieval can partially match the disease area while missing critical details, such as whether the setting is metastatic or adjuvant or the exact comparator regimen. These findings suggest that agentic retrieval alone is insufficient for MedKIT: successful performance requires not only locating topically relevant documents, but correctly aligning disease, treatment combination, clinical setting, endpoint, and comparative direction.

\subsection{Refusal Rates}
\label{app:refusal_rates}

\begin{figure}[h]
  \centering
  \includegraphics[width=0.85\linewidth]{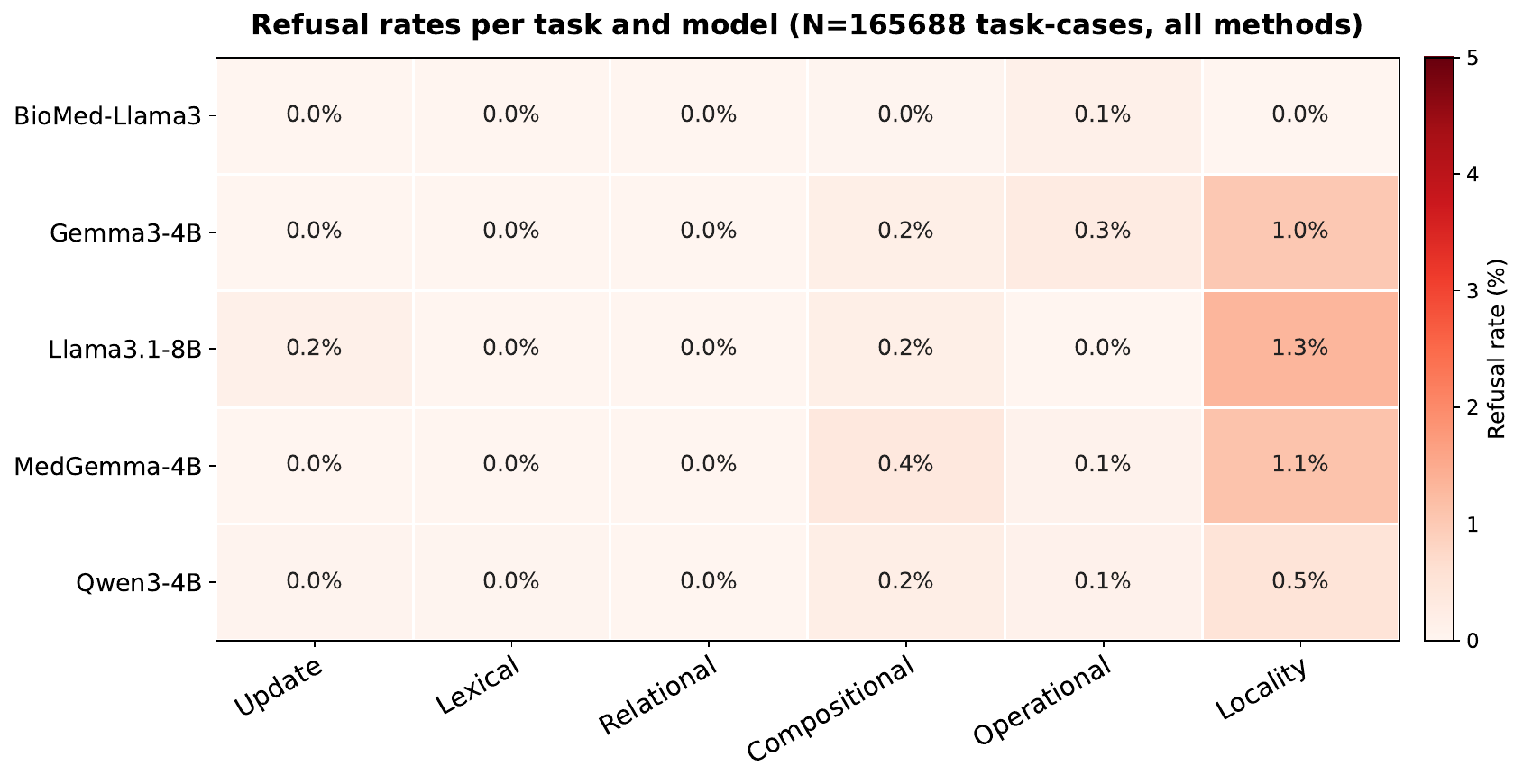}
  \caption{Refusal rates by task and model across all evaluated task cases.}
  \label{fig:refusal_rates}
\end{figure}

Figure~\ref{fig:refusal_rates} reports refusal rates across tasks and models. Refusals are rare across the full evaluation set and therefore do not explain the main performance trends. They occur most often on locality tasks, but remain below $1.5\%$ for all models.

\subsection{Capability Preservation}
\label{app:capability_preservation}

\begin{figure}[t]
  \centering
  \includegraphics[width=\textwidth]{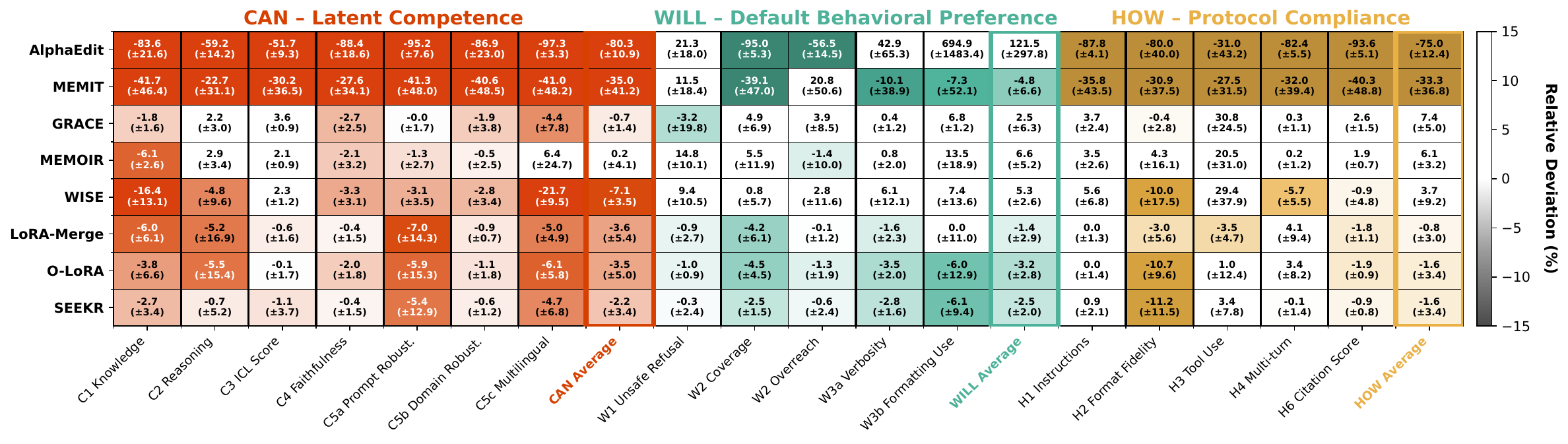}
  \caption{CapTrack relative deviation (\%) from the unedited base
    model, averaged across the five base models. Columns correspond
    to the full set of CAN / WILL / HOW capabilities together with
    per-category averages. Long-context tasks were not evaluated
    and therefore appear as $0$ by construction.}
  \label{fig:captrack_by_method}
\end{figure}

To complement the aggregate CapTrack analysis in \S\ref{app:capability_preservation}, Figure~\ref{fig:captrack_by_method} reports the full capability-level breakdown for all evaluated methods, averaged across the five base models. CapTrack measures relative deviation (\%) from the base model across three categories: \emph{latent competence} (CAN), \emph{behavioral preferences} (WILL), and \emph{protocol compliance} (HOW).

\paragraph{Setup.}
Two scope reductions apply to the runs reported here. First, long-context tasks were not evaluated for any method due to computational constraints. Second, the augmented editing methods \textbf{GRACE}, \textbf{WISE}, and \textbf{MEMOIR} were evaluated on a subsampled CapTrack split to reduce runtime. Their capability estimates are therefore noisier than those of the parameter-editing and continual fine-tuning methods, but still sufficient to assess the direction and scale of capability drift.

\paragraph{Parameter editing.}
\textbf{AlphaEdit} and \textbf{MEMIT} exhibit the strongest out-of-domain degradation. The largest drops occur in CAN capabilities related to factual knowledge, reasoning robustness, and multilingual robustness, together with substantial HOW degradation in instruction following, format fidelity, multi-turn interaction, and citation behavior. AlphaEdit shows the most severe drift overall, while MEMIT exhibits the same qualitative pattern at smaller magnitude. These results align with the within-domain locality findings in \S\ref{app:capability_preservation}: large weight updates that strongly integrate new knowledge also disrupt general capabilities outside the editing domain.

\paragraph{Augmented editing.}
\textbf{GRACE}, \textbf{MEMOIR}, and \textbf{WISE} remain comparatively stable across all three capability groups. Most deviations stay close to the unedited baseline, with no systematic collapse in CAN, WILL, or HOW capabilities. This stability is expected, as these methods primarily rely on external memories, routing modules, or retrieval-style mechanisms rather than directly modifying the base weights.

\paragraph{Continual fine-tuning.}
\textbf{LoRA-Merge}, \textbf{O-LoRA} and \textbf{SEEKR} produce moderate but substantially smaller drift than parameter editing. The strongest effects again appear in HOW-related capabilities, particularly instruction following and formatting behavior, while CAN capabilities remain largely preserved.

\paragraph{Retrieval augmentation.}
As retrieval-based methods do not alter the model weights or architecture, they have no impact on the general language model capabilities.

\paragraph{Summary.}
Overall, the full CapTrack breakdown reinforces the main finding of \S\ref{subsec:experiments_preservation}: methods that aggressively modify model weights achieve stronger immediate integration but at the cost of broader capability degradation, whereas augmented editing and low-rank continual fine-tuning preserve general model behavior substantially better.

\subsection{Complete Numerical Results}
\label{app:complete_results}

For completeness and reproducibility, Tables~\ref{tab:complete_generalization}--\ref{tab:complete_preservation} report the numerical results underlying Figures~\ref{fig:generalization_combined}--\ref{fig:capability_preservation}. We provide results separately for each method, model, and task type rather than only the aggregate values shown in the main text. All values follow the evaluation protocol described in \S\ref{app:evaluation_protocol} and report post--pre performance changes unless stated otherwise.

\begin{longtable}{llrrrrr}
\caption{Complete per-(method, model) generalization results underlying Figure~\ref{fig:generalization_combined}. Post--pre performance change (percentage points) per task tier; the two open-ended tiers (Compositional, Operational) are normalized to $[0,1]$ before differencing. Oracle-retrieval rows are an upper-bound reference not drawn in Figure~\ref{fig:generalization_combined}.}
\label{tab:complete_generalization} \\

\toprule
Method & Model & Update & Lexical & Relational & Compositional & Operational \\
\midrule
\endfirsthead

\multicolumn{7}{c}{\tablename\ \thetable{} -- continued from previous page} \\
\toprule
Method & Model & Update & Lexical & Relational & Compositional & Operational \\
\midrule
\endhead

\midrule
\multicolumn{7}{r}{\emph{Continued on next page}} \\
\endfoot

\bottomrule
\endlastfoot

\addlinespace
\multicolumn{7}{l}{\emph{Parameter editing}} \\
AlphaEdit & Llama-3.1-8B & $+9.1$ & $-1.9$ & $-7.4$ & $-67.8$ & $-28.3$ \\
 & Qwen-3-4B & $-15.6$ & $-19.3$ & $-21.5$ & $-72.6$ & $-37.6$ \\
 & Gemma-3-4B & $-16.0$ & $-23.1$ & $-15.6$ & $-42.6$ & $+0.6$ \\
 & MedGemma-4B & $-29.1$ & $-38.4$ & $-37.0$ & $-61.6$ & $-5.1$ \\
 & Bio-Med-Llama-3-8B & $-0.0$ & $-1.8$ & $-7.1$ & $-49.6$ & $-35.4$ \\
MEMIT & Llama-3.1-8B & $-8.4$ & $-12.6$ & $-18.2$ & $-66.3$ & $-29.9$ \\
 & Qwen-3-4B & $+45.4$ & $+36.4$ & $+7.0$ & $-72.6$ & $-17.2$ \\
 & Gemma-3-4B & $+29.8$ & $+13.2$ & $-2.3$ & $-70.3$ & $-9.5$ \\
 & MedGemma-4B & $+28.8$ & $+19.2$ & $-5.3$ & $-40.3$ & $-0.7$ \\
 & Bio-Med-Llama-3-8B & $+8.6$ & $+1.6$ & $-7.3$ & $-47.3$ & $-35.2$ \\
\addlinespace
\multicolumn{7}{l}{\emph{Augmented editing}} \\
GRACE & Llama-3.1-8B & $+59.4$ & $+1.8$ & $+0.3$ & $-0.7$ & $+0.6$ \\
 & Qwen-3-4B & $+65.1$ & $+0.1$ & $+0.4$ & $+0.2$ & $-0.5$ \\
 & Gemma-3-4B & $+45.0$ & $-0.3$ & $+0.6$ & $-0.4$ & $+0.8$ \\
 & MedGemma-4B & $+55.8$ & $+0.6$ & $+1.3$ & $+2.2$ & $+0.9$ \\
 & Bio-Med-Llama-3-8B & $+4.8$ & $+0.4$ & $+0.2$ & $+0.6$ & $+0.5$ \\
MEMOIR & Llama-3.1-8B & $+57.7$ & $+51.5$ & $+13.7$ & $-1.6$ & $-1.1$ \\
 & Qwen-3-4B & $+66.5$ & $+53.6$ & $+9.5$ & $+2.0$ & $-0.3$ \\
 & Gemma-3-4B & $+62.9$ & $+50.6$ & $+13.6$ & $-6.4$ & $-1.4$ \\
 & MedGemma-4B & $+63.4$ & $+54.9$ & $+16.5$ & $-1.9$ & $-3.6$ \\
 & Bio-Med-Llama-3-8B & $+49.9$ & $+45.3$ & $+16.1$ & $+0.9$ & $-2.2$ \\
WISE & Llama-3.1-8B & $+14.0$ & $+4.3$ & $-2.5$ & $-4.6$ & $+0.6$ \\
 & Qwen-3-4B & $+7.5$ & $+0.9$ & $-0.2$ & $-1.5$ & $-3.2$ \\
 & Gemma-3-4B & $+8.4$ & $+1.3$ & $+1.6$ & $-3.2$ & $-0.7$ \\
 & MedGemma-4B & $+19.4$ & $+11.6$ & $-0.4$ & $+1.7$ & $-1.6$ \\
 & Bio-Med-Llama-3-8B & $+23.1$ & $+18.0$ & $-1.2$ & $-5.7$ & $-4.7$ \\
IKE & Llama-3.1-8B & $+50.3$ & $-1.0$ & $-0.7$ & $+1.4$ & $-0.0$ \\
 & Qwen-3-4B & $+46.8$ & $-0.1$ & $+0.0$ & $+1.4$ & $-1.1$ \\
 & Gemma-3-4B & $+40.1$ & $+0.5$ & $+1.1$ & $-0.1$ & $+0.8$ \\
 & MedGemma-4B & $+55.2$ & $-2.4$ & $-3.2$ & $+2.1$ & $-0.1$ \\
 & Bio-Med-Llama-3-8B & $+61.8$ & $-0.1$ & $-0.1$ & $+0.0$ & $+0.4$ \\
\addlinespace
\multicolumn{7}{l}{\emph{Continual fine-tuning}} \\
LoRA-Merge & Llama-3.1-8B & $+65.0$ & $+54.8$ & $+31.0$ & $+2.3$ & $+1.1$ \\
 & Qwen-3-4B & $+52.0$ & $+38.3$ & $+19.7$ & $+2.4$ & $-3.5$ \\
 & Gemma-3-4B & $+38.1$ & $+21.6$ & $+9.7$ & $+1.5$ & $-0.1$ \\
 & MedGemma-4B & $+47.3$ & $+31.1$ & $+18.8$ & $-0.1$ & $-0.6$ \\
 & Bio-Med-Llama-3-8B & $+53.4$ & $+47.0$ & $+20.6$ & $-21.7$ & $-0.6$ \\
O-LoRA & Llama-3.1-8B & $+65.4$ & $+50.9$ & $+30.6$ & $+3.4$ & $+1.0$ \\
 & Qwen-3-4B & $+67.9$ & $+54.7$ & $+28.9$ & $+4.6$ & $-3.0$ \\
 & Gemma-3-4B & $+63.6$ & $+48.1$ & $+21.0$ & $+1.8$ & $+1.0$ \\
 & MedGemma-4B & $+63.7$ & $+52.4$ & $+24.3$ & $+1.3$ & $+0.0$ \\
 & Bio-Med-Llama-3-8B & $+52.3$ & $+45.6$ & $+19.6$ & $-21.0$ & $-1.2$ \\
SEEKR & Llama-3.1-8B & $+66.1$ & $+52.5$ & $+31.7$ & $+2.6$ & $+1.9$ \\
 & Qwen-3-4B & $+68.9$ & $+57.9$ & $+29.2$ & $+4.5$ & $-1.3$ \\
 & Gemma-3-4B & $+63.2$ & $+44.5$ & $+25.0$ & $-2.1$ & $-0.1$ \\
 & MedGemma-4B & $+63.4$ & $+51.9$ & $+24.9$ & $+1.2$ & $-0.3$ \\
 & Bio-Med-Llama-3-8B & $+60.1$ & $+52.9$ & $+26.6$ & $-8.8$ & $-2.1$ \\
\addlinespace
\multicolumn{7}{l}{\emph{Retrieval augmentation}} \\
BM25 RAG & Llama-3.1-8B & $+10.8$ & $-0.5$ & $-0.7$ & $+1.2$ & $+1.0$ \\
 & Qwen-3-4B & $+12.6$ & $+0.1$ & $+0.0$ & $-0.3$ & $-1.3$ \\
 & Gemma-3-4B & $+12.2$ & $-1.0$ & $+0.6$ & $+1.6$ & $-0.3$ \\
 & MedGemma-4B & $+10.5$ & $-0.2$ & $+0.7$ & $+1.6$ & $+0.7$ \\
 & Bio-Med-Llama-3-8B & $+9.7$ & $-1.4$ & $-0.8$ & $+0.1$ & $+0.5$ \\
Dense RAG & Llama-3.1-8B & $-0.4$ & $-0.1$ & $-1.1$ & $-1.3$ & $-1.2$ \\
 & Qwen-3-4B & $-0.8$ & $-0.1$ & $+0.0$ & $+0.2$ & $+1.0$ \\
 & Gemma-3-4B & $-1.2$ & $-0.7$ & $+0.6$ & $+0.6$ & $+0.5$ \\
 & MedGemma-4B & $+0.7$ & $-0.0$ & $+0.3$ & $+1.5$ & $-0.1$ \\
 & Bio-Med-Llama-3-8B & $+0.9$ & $-0.3$ & $+0.9$ & $+1.1$ & $+0.8$ \\
\midrule
\multicolumn{7}{l}{\emph{Reference: oracle retrieval (upper bound; not shown in Fig.~\ref{fig:generalization_combined})}} \\
Oracle (GT) & Llama-3.1-8B & $+65.0$ & $+59.6$ & $+19.7$ & $+21.5$ & $+49.2$ \\
 & Qwen-3-4B & $+52.7$ & $+46.8$ & $+22.9$ & $+7.5$ & $+23.6$ \\
 & Gemma-3-4B & $+58.7$ & $+50.7$ & $+14.6$ & $+14.7$ & $+42.3$ \\
 & MedGemma-4B & $+60.9$ & $+53.7$ & $+28.1$ & $+26.2$ & $+32.9$ \\
 & Bio-Med-Llama-3-8B & $+60.8$ & $+55.3$ & $-4.7$ & $+28.7$ & $+20.1$ \\
Oracle (Abs) & Llama-3.1-8B & $+29.5$ & $+20.1$ & $+26.7$ & $+19.8$ & $+47.3$ \\
 & Qwen-3-4B & $+29.2$ & $+23.2$ & $+23.9$ & $+13.7$ & $+37.1$ \\
 & Gemma-3-4B & $+27.7$ & $+17.3$ & $+22.4$ & $+10.5$ & $+43.4$ \\
 & MedGemma-4B & $+27.4$ & $+15.5$ & $+22.5$ & $+20.9$ & $+35.6$ \\
 & Bio-Med-Llama-3-8B & $+24.2$ & $+18.9$ & $+21.7$ & $+29.0$ & $+35.8$ \\
\end{longtable}

\begin{longtable}{llrrr}
\caption{Complete per-(method, model) sequential-retention results underlying Figure~\ref{fig:past_performance}. Closed-QA accuracy change (percentage points) vs.\ the per-model pre-edit baseline: \emph{Current}~$\Delta$ on the just-integrated update, \emph{Previous}~$\Delta$ on previously integrated updates (sentinel probe), and \emph{Drop}~$=$~Current~$-$~Previous.}
\label{tab:complete_retention} \\

\toprule
Method & Model & Current $\Delta$ & Previous $\Delta$ & Drop \\
\midrule
\endfirsthead

\multicolumn{5}{c}{\tablename\ \thetable{} -- continued from previous page} \\
\toprule
Method & Model & Current $\Delta$ & Previous $\Delta$ & Drop \\
\midrule
\endhead

\midrule
\multicolumn{5}{r}{\emph{Continued on next page}} \\
\endfoot

\bottomrule
\endlastfoot

\addlinespace
\multicolumn{5}{l}{\emph{Parameter editing}} \\
AlphaEdit & Llama-3.1-8B & $+0.8$ & $-11.9$ & $+12.8$ \\
 & Qwen-3-4B & $-18.1$ & $-21.5$ & $+3.3$ \\
 & Gemma-3-4B & $-18.9$ & $-17.4$ & $-1.5$ \\
 & MedGemma-4B & $-34.5$ & $-32.9$ & $-1.6$ \\
 & Bio-Med-Llama-3-8B & $-1.6$ & $-9.9$ & $+8.3$ \\
MEMIT & Llama-3.1-8B & $-11.4$ & $-20.3$ & $+8.9$ \\
 & Qwen-3-4B & $+33.5$ & $+7.2$ & $+26.3$ \\
 & Gemma-3-4B & $+15.1$ & $+5.6$ & $+9.5$ \\
 & MedGemma-4B & $+17.7$ & $+1.4$ & $+16.3$ \\
 & Bio-Med-Llama-3-8B & $+2.2$ & $-10.5$ & $+12.7$ \\
\addlinespace
\multicolumn{5}{l}{\emph{Augmented editing}} \\
GRACE & Llama-3.1-8B & $+14.6$ & $+16.6$ & $-2.0$ \\
 & Qwen-3-4B & $+14.0$ & $+16.3$ & $-2.2$ \\
 & Gemma-3-4B & $+10.0$ & $+11.7$ & $-1.7$ \\
 & MedGemma-4B & $+13.6$ & $+17.0$ & $-3.5$ \\
 & Bio-Med-Llama-3-8B & $+2.3$ & $+5.1$ & $-2.8$ \\
MEMOIR & Llama-3.1-8B & $+46.5$ & $+14.4$ & $+32.1$ \\
 & Qwen-3-4B & $+48.4$ & $+15.6$ & $+32.9$ \\
 & Gemma-3-4B & $+47.1$ & $+15.0$ & $+32.1$ \\
 & MedGemma-4B & $+50.6$ & $+23.0$ & $+27.6$ \\
 & Bio-Med-Llama-3-8B & $+41.3$ & $+15.6$ & $+25.7$ \\
WISE & Llama-3.1-8B & $+6.5$ & $-1.8$ & $+8.3$ \\
 & Qwen-3-4B & $+3.0$ & $+0.3$ & $+2.7$ \\
 & Gemma-3-4B & $+4.0$ & $+3.1$ & $+0.9$ \\
 & MedGemma-4B & $+12.3$ & $-0.3$ & $+12.7$ \\
 & Bio-Med-Llama-3-8B & $+16.1$ & $-3.1$ & $+19.2$ \\
IKE & Llama-3.1-8B & $+16.0$ & $+3.5$ & $+12.6$ \\
 & Qwen-3-4B & $+18.5$ & $+1.3$ & $+17.2$ \\
 & Gemma-3-4B & $+17.3$ & $-0.8$ & $+18.0$ \\
 & MedGemma-4B & $+20.1$ & $+0.8$ & $+19.3$ \\
 & Bio-Med-Llama-3-8B & $+24.6$ & $+5.2$ & $+19.4$ \\
\addlinespace
\multicolumn{5}{l}{\emph{Continual fine-tuning}} \\
LoRA-Merge & Llama-3.1-8B & $+53.7$ & $+25.4$ & $+28.2$ \\
 & Qwen-3-4B & $+38.1$ & $+17.6$ & $+20.5$ \\
 & Gemma-3-4B & $+23.6$ & $+13.1$ & $+10.6$ \\
 & MedGemma-4B & $+34.1$ & $+9.5$ & $+24.6$ \\
 & Bio-Med-Llama-3-8B & $+44.2$ & $+12.2$ & $+32.0$ \\
O-LoRA & Llama-3.1-8B & $+51.5$ & $+25.2$ & $+26.3$ \\
 & Qwen-3-4B & $+52.8$ & $+27.1$ & $+25.8$ \\
 & Gemma-3-4B & $+47.5$ & $+20.7$ & $+26.8$ \\
 & MedGemma-4B & $+50.8$ & $+18.5$ & $+32.3$ \\
 & Bio-Med-Llama-3-8B & $+42.9$ & $+13.0$ & $+29.9$ \\
SEEKR & Llama-3.1-8B & $+52.9$ & $+37.1$ & $+15.7$ \\
 & Qwen-3-4B & $+55.0$ & $+39.5$ & $+15.5$ \\
 & Gemma-3-4B & $+45.9$ & $+21.2$ & $+24.7$ \\
 & MedGemma-4B & $+50.9$ & $+24.1$ & $+26.8$ \\
 & Bio-Med-Llama-3-8B & $+50.1$ & $+29.7$ & $+20.4$ \\
\addlinespace
\multicolumn{5}{l}{\emph{Retrieval augmentation}} \\
BM25 RAG & Llama-3.1-8B & $+4.2$ & $+2.5$ & $+1.7$ \\
 & Qwen-3-4B & $+4.7$ & $+1.6$ & $+3.2$ \\
 & Gemma-3-4B & $+4.8$ & $-0.3$ & $+5.1$ \\
 & MedGemma-4B & $+4.3$ & $-0.2$ & $+4.5$ \\
 & Bio-Med-Llama-3-8B & $+3.1$ & $+4.4$ & $-1.3$ \\
Dense RAG & Llama-3.1-8B & $-0.3$ & $+2.7$ & $-3.0$ \\
 & Qwen-3-4B & $-0.2$ & $+1.7$ & $-1.9$ \\
 & Gemma-3-4B & $-0.7$ & $-0.7$ & $-0.1$ \\
 & MedGemma-4B & $+0.1$ & $+0.0$ & $+0.0$ \\
 & Bio-Med-Llama-3-8B & $+0.3$ & $+4.3$ & $-4.0$ \\
\end{longtable}

\begin{longtable}{llrr}
\caption{Complete per-(method, model) update-integration and within-domain locality results underlying Figure~\ref{fig:capability_preservation}(a). Post--pre $\Delta$ (percentage points); a large positive Update with near-zero Locality indicates the target update was integrated without damaging neighboring oncology facts.}
\label{tab:complete_preservation} \\

\toprule
Method & Model & Update $\Delta$ & Locality $\Delta$ \\
\midrule
\endfirsthead

\multicolumn{4}{c}{\tablename\ \thetable{} -- continued from previous page} \\
\toprule
Method & Model & Update $\Delta$ & Locality $\Delta$ \\
\midrule
\endhead

\midrule
\multicolumn{4}{r}{\emph{Continued on next page}} \\
\endfoot

\bottomrule
\endlastfoot

\addlinespace
\multicolumn{4}{l}{\emph{Parameter editing}} \\
AlphaEdit & Llama-3.1-8B & $+9.1$ & $-47.7$ \\
 & Qwen-3-4B & $-15.6$ & $-39.4$ \\
 & Gemma-3-4B & $-16.0$ & $-20.1$ \\
 & MedGemma-4B & $-29.1$ & $-52.7$ \\
 & Bio-Med-Llama-3-8B & $-0.0$ & $-29.2$ \\
MEMIT & Llama-3.1-8B & $-8.4$ & $-56.3$ \\
 & Qwen-3-4B & $+45.4$ & $-3.8$ \\
 & Gemma-3-4B & $+29.8$ & $-20.6$ \\
 & MedGemma-4B & $+28.8$ & $-22.0$ \\
 & Bio-Med-Llama-3-8B & $+8.6$ & $-33.8$ \\
\addlinespace
\multicolumn{4}{l}{\emph{Augmented editing}} \\
GRACE & Llama-3.1-8B & $+59.4$ & $+0.1$ \\
 & Qwen-3-4B & $+65.1$ & $-0.3$ \\
 & Gemma-3-4B & $+45.0$ & $-1.1$ \\
 & MedGemma-4B & $+55.8$ & $-0.5$ \\
 & Bio-Med-Llama-3-8B & $+4.8$ & $+0.0$ \\
MEMOIR & Llama-3.1-8B & $+57.7$ & $-0.8$ \\
 & Qwen-3-4B & $+66.5$ & $-0.0$ \\
 & Gemma-3-4B & $+62.9$ & $-1.7$ \\
 & MedGemma-4B & $+63.4$ & $+0.5$ \\
 & Bio-Med-Llama-3-8B & $+49.9$ & $+10.4$ \\
WISE & Llama-3.1-8B & $+14.0$ & $-1.4$ \\
 & Qwen-3-4B & $+7.5$ & $-1.4$ \\
 & Gemma-3-4B & $+8.4$ & $-1.1$ \\
 & MedGemma-4B & $+19.4$ & $+1.2$ \\
 & Bio-Med-Llama-3-8B & $+23.1$ & $+4.3$ \\
IKE & Llama-3.1-8B & $+50.3$ & $-0.3$ \\
 & Qwen-3-4B & $+46.8$ & $-0.4$ \\
 & Gemma-3-4B & $+40.1$ & $+2.6$ \\
 & MedGemma-4B & $+55.2$ & $+2.3$ \\
 & Bio-Med-Llama-3-8B & $+61.8$ & $+2.0$ \\
\addlinespace
\multicolumn{4}{l}{\emph{Continual fine-tuning}} \\
LoRA-Merge & Llama-3.1-8B & $+65.0$ & $-1.1$ \\
 & Qwen-3-4B & $+52.0$ & $+0.3$ \\
 & Gemma-3-4B & $+38.1$ & $-0.7$ \\
 & MedGemma-4B & $+47.3$ & $+0.7$ \\
 & Bio-Med-Llama-3-8B & $+53.4$ & $-31.1$ \\
O-LoRA & Llama-3.1-8B & $+65.4$ & $-1.4$ \\
 & Qwen-3-4B & $+67.9$ & $+0.1$ \\
 & Gemma-3-4B & $+63.6$ & $+0.0$ \\
 & MedGemma-4B & $+63.7$ & $+0.1$ \\
 & Bio-Med-Llama-3-8B & $+52.3$ & $-30.4$ \\
SEEKR & Llama-3.1-8B & $+66.1$ & $-1.0$ \\
 & Qwen-3-4B & $+68.9$ & $+0.1$ \\
 & Gemma-3-4B & $+63.2$ & $+0.2$ \\
 & MedGemma-4B & $+63.4$ & $+0.7$ \\
 & Bio-Med-Llama-3-8B & $+60.1$ & $-21.2$ \\
\addlinespace
\multicolumn{4}{l}{\emph{Retrieval augmentation}} \\
BM25 RAG & Llama-3.1-8B & $+10.8$ & $-0.3$ \\
 & Qwen-3-4B & $+12.6$ & $+0.2$ \\
 & Gemma-3-4B & $+12.2$ & $-0.8$ \\
 & MedGemma-4B & $+10.5$ & $-0.6$ \\
 & Bio-Med-Llama-3-8B & $+9.7$ & $-0.5$ \\
Dense RAG & Llama-3.1-8B & $-0.4$ & $-0.0$ \\
 & Qwen-3-4B & $-0.8$ & $+0.5$ \\
 & Gemma-3-4B & $-1.2$ & $-0.9$ \\
 & MedGemma-4B & $+0.7$ & $-0.0$ \\
 & Bio-Med-Llama-3-8B & $+0.9$ & $-0.8$ \\
\end{longtable}

\begin{longtable}{lrrr}
\caption{Complete out-of-domain capability \emph{forgetting} results underlying Figure~\ref{fig:capability_preservation}(b). CapTrack relative deviation magnitude (\%) vs.\ the unedited base model, averaged over models; larger is worse. Only degradations count (per-probe improvements are clipped to $0$ before averaging), and weight-preserving methods (IKE, RAG variants) are $0$ by construction.}
\label{tab:complete_captrack} \\

\toprule
Method & Latent Competence & Behavioural Preferences & Protocol Compliance \\
\midrule
\endfirsthead

\multicolumn{4}{c}{\tablename\ \thetable{} -- continued from previous page} \\
\toprule
Method & Latent Competence & Behavioural Preferences & Protocol Compliance \\
\midrule
\endhead

\midrule
\multicolumn{4}{r}{\emph{Continued on next page}} \\
\endfoot

\bottomrule
\endlastfoot

\addlinespace
\multicolumn{4}{l}{\emph{Parameter editing}} \\
AlphaEdit & $78.7$ & $52.7$ & $82.3$ \\
MEMIT & $35.5$ & $26.0$ & $39.4$ \\
\addlinespace
\multicolumn{4}{l}{\emph{Augmented editing}} \\
GRACE & $3.6$ & $3.9$ & $1.8$ \\
MEMOIR & $4.6$ & $3.5$ & $5.0$ \\
WISE & $10.2$ & $5.0$ & $11.2$ \\
IKE & $0.0$ & $0.0$ & $0.0$ \\
\addlinespace
\multicolumn{4}{l}{\emph{Continual fine-tuning}} \\
LoRA-Merge & $5.3$ & $3.8$ & $3.5$ \\
O-LoRA & $5.1$ & $5.3$ & $5.6$ \\
SEEKR & $3.5$ & $4.1$ & $4.9$ \\
\addlinespace
\multicolumn{4}{l}{\emph{Retrieval augmentation}} \\
BM25 RAG & $0.0$ & $0.0$ & $0.0$ \\
Dense RAG & $0.0$ & $0.0$ & $0.0$ \\
Agentic RAG & $0.0$ & $0.0$ & $0.0$ \\
\end{longtable}

\subsection{Inference Cost at Deployment}
\label{app:deployment_cost}

In addition to update quality, practical deployment of knowledge integration methods depends on both the cost of incorporating new evidence and the cost of serving the resulting model. We therefore measure two complementary quantities for the weekly update setting: (i) \emph{edit time per update}, computed from the average adaptation time reported by each method during the main sweep, and (ii) \emph{inference time per case}, measured on the final post-edit checkpoint after all updates have been integrated.

\begin{figure}[t]
    \centering
    \includegraphics[width=0.7\linewidth]{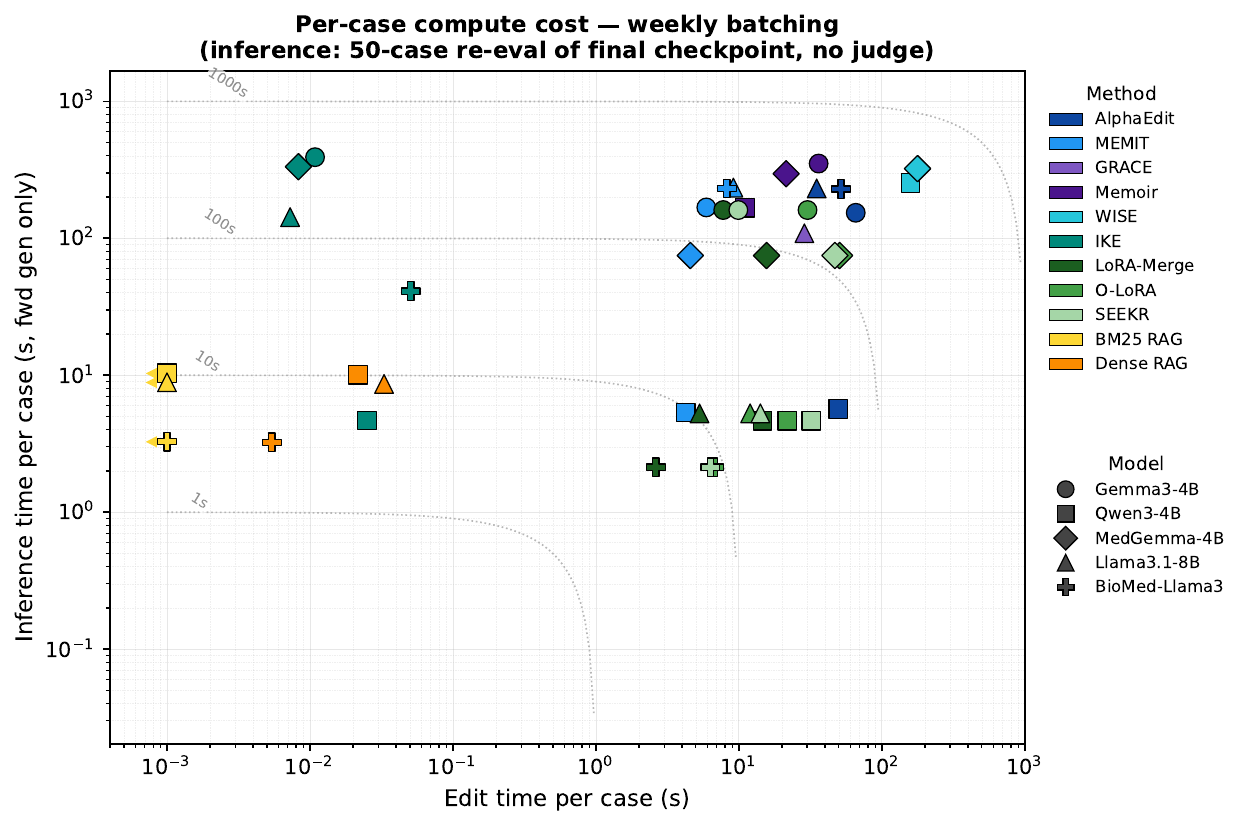}
    \caption{
    Deployment cost of knowledge integration methods under the weekly update setting. Each point corresponds to a method--model pair, showing average edit time per update (x-axis, log scale) against average inference time per evaluation case (y-axis, log scale). Two clear inference regimes emerge: methods compatible with \texttt{vLLM} achieve substantially lower deployment-time inference cost, while methods requiring custom HuggingFace forward passes incur one to two orders of magnitude slower generation. Retrieval methods exhibit negligible edit cost but remain constrained by inference-time retrieval and prompt length overheads.
    }
    \label{fig:deployment_cost_scatter}
\end{figure}

For inference measurements, we evaluate a fixed set of 50 post-2025 benchmark cases using the same eight evaluation probes employed in the main experiments (anchor, lexical, relational, five portability variants, and locality), excluding the LLM judge. We report the average wall-clock time required to generate all responses for one case under batched generation. Figure~\ref{fig:deployment_cost_scatter} summarizes the resulting edit--vs--inference trade-off across methods and models.

\paragraph{Backend compatibility dominates inference cost.}
The primary determinant of inference efficiency is whether the edited model can be served through \texttt{vLLM}. Methods that produce a standard HuggingFace \texttt{state\_dict} after editing, including MEMIT, AlphaEdit, the continual fine-tuning methods (LoRA-Merge, O-LoRA, SEEKR), IKE, and the retrieval baselines, can be evaluated using \texttt{vLLM} with efficient batched decoding. In contrast, GRACE, Memoir, and WISE modify the forward pass through custom routing or side-network mechanisms that are not supported by \texttt{vLLM}, requiring substantially slower HuggingFace-based generation.

This distinction produces a clear two-band structure in Figure~\ref{fig:deployment_cost_scatter}. \texttt{vLLM}-compatible methods achieve inference times of approximately $2$--$10$ seconds per case, whereas HuggingFace-only methods require roughly $100$--$500$ seconds per case depending on model size. Retrieval-augmented approaches appear at the upper end of the fast band due to longer augmented prompts, while all Gemma-family measurements fall into the slower regime because the Gemma-3 multimodal wrapper currently prevents stable \texttt{vLLM} initialization in our setup.

\paragraph{Inference dominates deployment cost.}
Edit cost spans several orders of magnitude, ranging from effectively negligible retrieval-index updates to over $100$ seconds per update for side-network training methods such as WISE. However, for nearly all approaches, the cumulative deployment cost is dominated by repeated inference rather than by the one-time integration step. Even methods with relatively expensive updates incur substantially larger wall-clock cost during routine usage if they cannot be served efficiently.

Overall, our measurements suggest that deployment efficiency is determined more strongly by the serving backend supported by the edited artifact than by the nominal update cost of the editing method itself. Across models, serving through \texttt{vLLM} reduces inference cost by approximately an order of magnitude compared to HuggingFace-based execution, substantially outweighing most differences in per-update training cost.

\section{Limitations and Broader Considerations}
\label{app:limitations_ethics}

\subsection{Benchmark Limitations}
\label{app:benchmark_limitations}

\paragraph{Pairwise formulation.}
MedKIT represents each update as a pairwise comparison between two regimens on a single endpoint with labels $\{\textit{superior}, \textit{inferior}, \textit{no difference}\}$. This aligns with trial reporting and enables deterministic probe construction, but abstracts away clinically relevant structure. In particular, MedKIT evaluates whether models integrate and generalize the qualitative conclusion of a treatment comparison rather than quantitative treatment effects. Multi-arm relationships, effect sizes, uncertainty, and subgroup-specific effects are not explicitly modeled. As a result, MedKIT measures the integration of pairwise comparisons rather than the full clinical decision-making process. Extending the benchmark to quantitative evidence and uncertainty represents an important direction for future work.

\paragraph{Source-data biases.}
MedKIT inherits biases from HemOnc.org and the underlying clinical literature. Coverage across oncology groups is uneven, label assignments reflect curator judgment, and endpoint prioritization introduces normative assumptions. Additionally, PubMed-indexed trials over-represent Western populations. These factors affect the distribution of benchmark instances and limit generalizability beyond the source data.

\subsection{Evaluation Limitations}
\label{app:evaluation_limitations}

\paragraph{LLM judge.}
Open-form tasks rely on an LLM judge (\S\ref{app:judge_evaluation}). Agreement with the two-clinician consensus is high on compositional tasks but lower on operational generation (\S\ref{app:judge_validation}), reflecting the substantially broader space of acceptable responses in open-ended recommendation generation. The clinician study is limited to two expert annotators and 100 responses; broader multi-clinician evaluation would provide a more precise estimate of variability across clinical experts.

Importantly, MedKIT evaluates \emph{relative} changes in downstream behavior (e.g., post--pre deltas and comparative method performance) rather than absolute clinical quality scores. While fine-grained method rankings on open-form tasks are sensitive to small score fluctuations, the higher-level comparative conclusions used throughout the paper remain stable across judges, particularly at the level of method families and large performance differences (\S\ref{app:judge_validation}).

\paragraph{Open-generation scoring.}
The operational rubric evaluates \emph{consistency} with the underlying comparison, not clinical correctness. It does not verify dosage, contraindications, or safety considerations. Scores should therefore be interpreted as measuring knowledge use rather than clinical validity.

\paragraph{Refusals and stochasticity.}
Refusals are scored as incorrect (closed) or neutral (open), which does not distinguish between safe abstention and incorrect answers. Open-form scores depend on stochastic judge calls, introducing small variance despite generally stable rankings (\S\ref{app:judge_evaluation}).

\subsection{Medical and Deployment Considerations}
\label{app:medical_deployment}

\paragraph{Research benchmark.}
MedKIT is designed for method evaluation, not clinical decision support. Scores reflect the integration of benchmark-specific knowledge and do not imply clinical readiness.

\paragraph{Failure modes.}
Several observed failure modes are critical in practice:
(i) correct recall without correct use (open-generation mismatch),
(ii) hallucinated or degraded outputs after editing, and
(iii) locality spillover to unrelated medical contexts. These are
not detectable from closed-form evaluation alone.

\paragraph{Deployment implications.}
Any deployment setting requires additional safeguards beyond
benchmark performance. In particular: (i) validation on
clinician-written queries, (ii) continuous monitoring under
sequential updates, and (iii) preference for abstention over
potentially harmful recommendations.

\paragraph{Data and governance.}
The benchmark uses public data (HemOnc.org and PubMed) and contains
no patient-level information. Deployment scenarios involving sensitive clinical data introduce privacy and governance challenges that are outside the scope of MedKIT.

\paragraph{Potential positive impact.}
Because MedKIT is grounded in real-world clinical evidence and temporally structured updates, progress on the benchmark is more closely aligned with practical knowledge-maintenance challenges than progress on synthetic or purely recall-centric editing tasks. Methods that successfully integrate and apply evolving knowledge across the diverse settings captured by MedKIT 
could inform the development of methods for maintaining language models in domains where outdated information introduces meaningful risks.

\subsection{Licenses and Existing Assets}
\label{app:licenses}

\paragraph{Clinical source data.}
MedKIT is derived from HemOnc.org~\citep{warner2019hemonc}, a publicly accessible oncology knowledge base licensed under the CC BY-NC-SA 4.0 license for academic and non-commercial use. The benchmark additionally uses linked PubMed metadata and abstracts retrieved through publicly available NCBI services. We release only processed benchmark instances and evaluation artifacts necessary for research use and provide attribution to the original sources.

\paragraph{Models.}
Experiments use publicly released language models, including Gemma-3, MedGemma, Qwen-3, LLaMA-3.1, and Bio-Medical-LLaMA-3, subject to their respective licenses and acceptable-use policies.

\paragraph{Released assets.}
The MedKIT benchmark will be released under the CC BY-NC-SA 4.0 license, consistent with the licensing terms of the underlying HemOnc.org data. The evaluation code and experimental framework will be released under the Apache 2.0 license. The release additionally includes documentation covering dataset construction, preprocessing, evaluation protocols, data fields, and instructions for reproducing the main experiments.

\newpage
\section*{NeurIPS Paper Checklist}

\begin{enumerate}

\item {\bf Claims}
    \item[] Question: Do the main claims made in the abstract and introduction accurately reflect the paper's contributions and scope?
    \item[] Answer: \answerYes{}
    \item[] Justification: The abstract and introduction accurately summarize the benchmark construction, evaluation setting, and empirical findings. The paper clearly states its main contributions, including the MedKIT benchmark, the multi-level generalization framework, and the large-scale evaluation of knowledge integration methods (§1, §3). Experimental results supporting the main claims are reported in §4, while assumptions and limitations are discussed in §5 and Appendix~F.

\item {\bf Limitations}
    \item[] Question: Does the paper discuss the limitations of the work performed by the authors?
    \item[] Answer: \answerYes{}
    \item[] Justification: The paper includes a dedicated limitations discussion in §5 and Appendix~F. These sections discuss limitations of the benchmark construction and evaluation protocol, including the pairwise formulation of clinical updates, source-data biases, reliance on LLM judges, scalability constraints of some methods, and the restricted model scale studied in the experiments. The appendix further discusses deployment considerations, evaluation limitations, and broader implications for medical settings.

\item {\bf Theory assumptions and proofs}
    \item[] Question: For each theoretical result, does the paper provide the full set of assumptions and a complete (and correct) proof?
    \item[] Answer: \answerNA{}
    \item[] Justification: The paper does not present theoretical results, formal theorems, or mathematical proofs. The work is entirely empirical, focusing on benchmark construction and experimental evaluation of knowledge integration methods.

    \item {\bf Experimental result reproducibility}
    \item[] Question: Does the paper fully disclose all the information needed to reproduce the main experimental results of the paper to the extent that it affects the main claims and/or conclusions of the paper (regardless of whether the code and data are provided or not)?
    \item[] Answer: \answerYes{}
    \item[] Justification: The paper describes the benchmark construction pipeline, preprocessing decisions, task generation, evaluation protocol, model setup, batching strategies, and implementation details in §3--§4 and Appendices~A--E. The released benchmark and codebase include the data-processing pipeline, evaluation framework, prompt templates, and experiment configurations needed to reproduce the main results, together with details on compute resources, seeds, and inference settings (Appendix~B and Appendix~C.4).

\item {\bf Open access to data and code}
    \item[] Question: Does the paper provide open access to the data and code, with sufficient instructions to faithfully reproduce the main experimental results, as described in supplemental material?
    \item[] Answer: \answerYes{}
    \item[] Justification: The paper provides links to both the MedKIT benchmark and the corresponding evaluation codebase in §3. The released assets include the benchmark construction pipeline, preprocessing scripts, prompt templates, evaluation framework, and experiment configurations, together with documentation describing dataset preparation, evaluation protocols, and reproduction of the main experiments (Appendices~A--D).

\item {\bf Experimental setting/details}
    \item[] Question: Does the paper specify all the training and test details (e.g., data splits, hyperparameters, how they were chosen, type of optimizer) necessary to understand the results?
    \item[] Answer: \answerYes{}
    \item[] Justification: The paper specifies the benchmark construction pipeline, temporal batching strategy, evaluation protocol, model selection, training and inference setup, and hyperparameter tuning procedure in §3--§4 and Appendices~A--E. Additional implementation details, including compute setup, decoding settings, seeds, and backend configuration, are provided in Appendix~C.4, while prompt templates and judge protocols are included in Appendix~A.4 and Appendix~B.

\item {\bf Experiment statistical significance}
    \item[] Question: Does the paper report error bars suitably and correctly defined or other appropriate information about the statistical significance of the experiments?
    \item[] Answer: \answerYes{}
    \item[] Justification: The main figures report either individual model-level results or error bars around aggregate metrics (Figures~3--5). Appendix~B.1 explains the aggregation procedure and specifies that error bars indicate standard deviation across evaluated models unless otherwise stated. The paper does not rely on formal null-hypothesis significance testing; instead, conclusions are supported by consistent effect patterns observed across models, methods, and task tiers.

\item {\bf Experiments compute resources}
    \item[] Question: For each experiment, does the paper provide sufficient information on the computer resources (type of compute workers, memory, time of execution) needed to reproduce the experiments?
    \item[] Answer: \answerYes{}
    \item[] Justification: Appendix~C.4 reports the hardware and compute setup used for all experiments, including GPU type (NVIDIA A100 40GB/80GB), cluster setup, per-run GPU allocation, inference backends, and total compute requirements. The paper also reports the approximate total compute budget for the experimental sweep and hyperparameter tuning, along with details on reproducibility settings, such as seeds and decoding configurations.
    
\item {\bf Code of ethics}
    \item[] Question: Does the research conducted in the paper conform, in every respect, with the NeurIPS Code of Ethics \url{https://neurips.cc/public/EthicsGuidelines}?
    \item[] Answer: \answerYes{}
    \item[] Justification: The authors have reviewed the NeurIPS Code of Ethics and believe the research conforms to its principles. The work uses publicly available clinical evidence sources without patient-level data, discusses limitations and deployment risks explicitly (§5 and Appendix~F), and releases the benchmark and code under licenses consistent with the underlying data sources (Appendix~F).

\item {\bf Broader impacts}
    \item[] Question: Does the paper discuss both potential positive societal impacts and negative societal impacts of the work performed?
    \item[] Answer: \answerYes{}
    \item[] Justification: The paper discusses both positive and negative societal implications in §5 and Appendix~F. Positive impacts include enabling more reliable maintenance and evaluation of language models in dynamic, high-stakes domains such as medicine. The paper additionally discusses risks and limitations, including incorrect clinical recommendations, locality spillover, degradation of general capabilities after updates, misuse of benchmark performance as evidence of clinical readiness, and the need for safeguards and clinician oversight in deployment settings.
    
\item {\bf Safeguards}
    \item[] Question: Does the paper describe safeguards that have been put in place for responsible release of data or models that have a high risk for misuse (e.g., pre-trained language models, image generators, or scraped datasets)?
    \item[] Answer: \answerYes{}
    \item[] Justification: The paper discusses safeguards and responsible-use considerations in §5 and Appendix~F. MedKIT is explicitly positioned as a research benchmark rather than a clinical decision-support system, and the paper highlights risks associated with incorrect recommendations, locality spillover, and capability degradation after updates. The released benchmark contains no patient-level data and is distributed under licenses consistent with the underlying public data sources, together with guidance emphasizing that deployment requires clinician oversight, validation, and additional safety safeguards.

\item {\bf Licenses for existing assets}
    \item[] Question: Are the creators or original owners of assets (e.g., code, data, models), used in the paper, properly credited and are the license and terms of use explicitly mentioned and properly respected?
    \item[] Answer: \answerYes{}
    \item[] Justification: The paper credits all major external assets used in the benchmark construction and experiments, including HemOnc.org, PubMed, and the evaluated language models (§3 and References). Appendix~F explicitly documents the licenses and terms of use of the underlying data sources and states that the released benchmark and code follow licensing terms consistent with the original assets.

\item {\bf New assets}
    \item[] Question: Are new assets introduced in the paper well documented and is the documentation provided alongside the assets?
    \item[] Answer: \answerYes{}
    \item[] Justification: The paper introduces the MedKIT benchmark together with a public release of the dataset and evaluation framework (§3). The benchmark construction pipeline, preprocessing decisions, task templates, evaluation protocol, dataset statistics, licensing information, and implementation details are documented throughout Appendices~A--F, and the released assets include accompanying documentation describing dataset fields, preprocessing, evaluation, and experiment reproduction.

\item {\bf Crowdsourcing and research with human subjects}
    \item[] Question: For crowdsourcing experiments and research with human subjects, does the paper include the full text of instructions given to participants and screenshots, if applicable, as well as details about compensation (if any)? 
    \item[] Answer: \answerNA{}
    \item[] Justification: The paper does not involve crowdsourcing experiments or research with human subjects. All clinician input and human annotations used for judge validation were provided by coauthors as part of the research process; no external annotators, crowdsourcing platforms, or patient studies were involved.

\item {\bf Institutional review board (IRB) approvals or equivalent for research with human subjects}
    \item[] Question: Does the paper describe potential risks incurred by study participants, whether such risks were disclosed to the subjects, and whether Institutional Review Board (IRB) approvals (or an equivalent approval/review based on the requirements of your country or institution) were obtained?
    \item[] Answer: \answerNA{}
    \item[] Justification: The paper does not involve crowdsourcing or research with human subjects. All clinician input and annotations used for judge validation were provided internally by coauthors as part of the research process, without external participants, patient interaction, or collection of personal data.

\item {\bf Declaration of LLM usage}
    \item[] Question: Does the paper describe the usage of LLMs if it is an important, original, or non-standard component of the core methods in this research? Note that if the LLM is used only for writing, editing, or formatting purposes and does \emph{not} impact the core methodology, scientific rigor, or originality of the research, declaration is not required.
    \item[] Answer: \answerYes{}
    \item[] Justification: The paper explicitly describes the use of LLMs as a core component of the evaluation methodology. In particular, §3.4 and Appendix~B detail the use of LLM-based judges for evaluating open-form tasks, including prompt design, validation against clinician annotations, robustness analyses across judge models, and the normalization and aggregation procedures used for scoring.

\end{enumerate}

\end{document}